\documentclass{article}

\usepackage[preprint]{neurips_2026}

\usepackage[utf8]{inputenc}
\usepackage[T1]{fontenc}

\usepackage{amsmath, amssymb, amsfonts}
\usepackage{mathtools}

\usepackage{graphicx}
\usepackage{booktabs}
\usepackage{multirow}
\usepackage{nicefrac}
\usepackage{xcolor}
\usepackage{microtype}

\usepackage{hyperref}
\usepackage{url}

\newcommand{\dM}{d_{\mathrm{M}}}

\title{PhysEDA: Physics-Aware Learning Framework for Efficient EDA
       With Manhattan Distance Decay}

\author{%
  Zetao Yang \\
  School of Mathematics and Statistics \\
  Beijing Institute of Technology \\
  Beijing, China \\
  \texttt{1120220294@bit.edu.cn}
}

\begin{document}

\maketitle

\begin{abstract}
Electronic design automation (EDA) addresses placement, routing, timing
analysis, and power-integrity verification for integrated circuits.
Learning methods---attention (Transformer) and reinforcement learning
(RL)---have recently emerged on EDA tasks, yet face two common
bottlenecks: vanilla attention's quadratic complexity limits scaling,
and data-scarce models overfit statistical noise and amplify weak
long-range correlations against the underlying physics. We observe that
EDA tasks share a physical prior---pairwise electrical and routing
interactions decay exponentially along Manhattan distance---and integrate
it as a unified inductive bias into both architecture and training. We propose
\textbf{PhysEDA}, comprising two components: \textbf{Physics-Structured
Linear Attention} (PSLA) folds the separable Manhattan decay into the
linear-attention kernel as a multiplicative bias, reducing complexity from
quadratic to linear; \textbf{Potential-Based Reward Shaping} (PBRS)
constructs a physical potential from the same kernel, providing dense
reward signal under sparse RL while preserving the optimal policy via the
policy-invariance theorem. Across three EDA scenarios---decoupling-capacitor
placement, macro placement, and IR-drop prediction---PhysEDA improves
zero-shot cross-scale transfer by $56.8\%$ and achieves $14\times$ inference
speedup with $98.5\%$ memory savings on $100\times 100$ grids; PBRS adds another
$10.8\%$ in sparse-reward DPP.
\end{abstract}

\section{Introduction}
\label{sec:intro}

Modern VLSI integrates billions of transistors per chip, and physical design must trade off performance, power, area, timing, and signal integrity. Electronic design automation (EDA) is the methodology that translates a transistor-level circuit into a manufacturable layout, spanning placement, routing, timing analysis, and power-integrity verification. Although these tasks differ in objective, they share one physical regularity: pairwise electrical and routing interactions decay exponentially along Manhattan distance~\citep{swaminathan2007power,fan2000quantifying}. We use this regularity as a unified prior in both architecture and training. Classical analytical placers like ePlace~\citep{lu2015eplace} and DREAMPlace~\citep{lin2019dreamplace} solve each design from scratch, motivating learning methods that amortize across designs.

Three EDA scenarios have driven a representative line of recent learning methods. For \emph{macro placement}, \citet{mirhoseini2021graph} and AlphaChip~\citep{mirhoseini2024alphachip} first applied GNN+RL to TPU placement; MaskPlace~\citep{lai2022maskplace} added dense pixel-level rewards; ChiPFormer~\citep{lai2023chipformer} reformulated placement as offline RL on a Decision Transformer~\citep{chen2021decision}; LaMPlace~\citep{li2025lamplace} optimizes for post-routing timing. For \emph{decoupling-capacitor placement} (DPP), DevFormer~\citep{kim2023devformer} used a symmetric Transformer with relative position encoding for the data-scarce DPP setting. For \emph{IR-drop prediction}, PowerNet~\citep{xie2020powernet}, IREDGe~\citep{chhabria2021iredge}, and PDNNet~\citep{zhao2024pdnnet} progressively introduced CNN, image-translation, and GNN+CNN dual-branch designs.

While each method achieves SOTA on its own task, three bottlenecks persist. The first is solution scale: these models typically use vanilla attention with quadratic complexity, so memory becomes prohibitive beyond $L\!\approx\!10{,}000$ candidate positions and blocks scaling to large designs. The second is data efficiency: implicitly recovering Manhattan-distance decay from data demands many samples, but chip-design data is naturally scarce---DevFormer's public DPP set has only 2{,}000 instances, and zero-shot cross-scale transfer provides no labels at all. Without the prior, models overfit and amplify spurious long-range correlations inconsistent with the underlying physics. The third is training efficiency: RL on EDA tasks suffers extreme reward sparsity---in DPP the simulator returns a non-zero reward only after all capacitors are placed, leaving every intermediate step unsupervised. Early training is close to random exploration, and convergence is slow. Underlying all three bottlenecks is a single root cause: training data is insufficient to recover spatial structure on its own. The same physical prior addresses all three.

\begin{figure*}[!t]
  \centering
  \includegraphics[width=\textwidth]{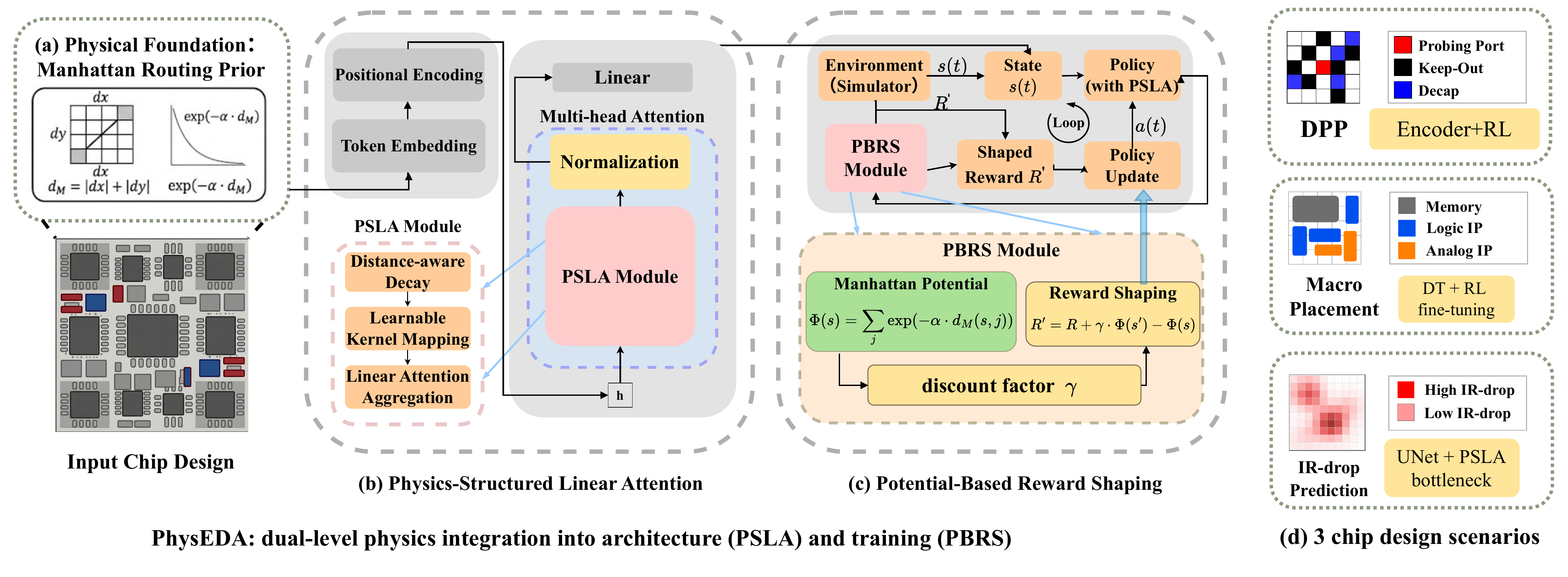}
  \caption{PhysEDA architecture. A single Manhattan-decay kernel (a) modulates linear attention as a learnable multiplicative bias (PSLA, b) and shapes the RL reward potential (PBRS, c), and is validated on three EDA tasks (d).}
  \label{fig:architecture}
\end{figure*}

We propose PhysEDA to address all three bottlenecks. Our central observation: the Manhattan decay $\exp(-\alpha\cdot\dM)$ is separable along $x$ and $y$, letting us integrate the same kernel at two levels of the pipeline. As a multiplicative bias inside the attention kernel, it yields \emph{Physics-Structured Linear Attention} (PSLA), which targets scale and data efficiency. As a potential function in reward shaping, it yields \emph{Potential-Based Reward Shaping} (PBRS), which targets training efficiency and preserves the optimal policy via the policy-invariance theorem~\citep{ng1999policy}. PhysEDA differs from prior work in letting one physical prior bridge architecture and training (Figure~\ref{fig:architecture}).

PhysEDA delivers three-fold contributions. At the architectural level, PSLA folds the separable Manhattan decay into linear attention as a learnable multiplicative bias, reducing complexity from quadratic to linear and delivering $14\times$ speedup with $98.5\%$ memory savings on $100\times 100$ grids ($32.5\times$ on $150\times 150$). At the training level, PBRS shares the same decay kernel to construct a physical potential, supplying dense shaping signal under sparse rewards while leaving the optimal policy unchanged, and on DPP $25\times 25$ RL it improves the score from $-15.70$ to $-17.40$ ($10.8\%$ relative). Empirically, we validate PhysEDA on three EDA scenarios with DPP supervised $+5.5\%$, DPP RL $+49.1\%$, DPP cross-scale transfer $+56.8\%$, macro placement HPWL $+12.0\%$ in Decision-Transformer pretraining with an additional $+5.4\%$ in RL fine-tuning, and IR-drop prediction cross-domain Pearson improvement of $+5.3\%$ to $+5.4\%$. Across all three scenarios, PhysEDA's gain monotonically tracks the data insufficiency for spatial structure, making the physical prior most valuable under data scarcity, distribution shift, and cross-scale transfer.

\section{The PhysEDA Framework}
\label{sec:framework}

PhysEDA comprises two components that share a single decay kernel $\exp(-\alpha\cdot \dM)$: PSLA (\S\ref{sec:psla}) encodes it as a multiplicative bias on the attention kernel, and PBRS (\S\ref{sec:pbrs}) encodes it as a reward potential. The two are complementary---PSLA shapes \emph{how} the model represents spatial relations, PBRS shapes \emph{how} training explores---and together they integrate the physical prior across architecture and training (\S\ref{sec:unified_block}).

\subsection{PSLA: Physics-Structured Linear Attention}
\label{sec:psla}

\subsubsection{Physical Motivation: From PDN to Attention Decay}
\label{sec:psla_motivation}

We show that the transfer impedance $Z_{\mathrm{tr}}(i,j,f)$---the voltage response at $i$ at frequency $f$ to a unit current injected at $j$---decays exponentially with Manhattan distance. Using the transmission-line mesh model~\citep{swaminathan2007power}, each mesh segment introduces a fixed voltage attenuation, so impedance over $n$ segments decays exponentially: $|Z_{\mathrm{tr}}|\propto\exp(-\kappa n)$.\footnote{The mesh model treats the PDN as a 2D grid of transmission-line segments (each with per-unit-length $R$, $L$, $C$); cavity-resonance models~\citep{kim2010chip} instead treat the parallel-plate PDN as a 2D resonant cavity. The two formulations agree numerically on standard PDN testcases.} With orthogonal routing~\citep{schulte2012design} and Manhattan distance as the dominant routing metric~\citep{koh2000manhattan_routing}, current from $j$ to $i$ traverses a staircase path of length $\dM\!\equiv\!|x_i\!-\!x_j|+|y_i\!-\!y_j|$, so $n=\dM$ yields, in the 100\,MHz--2\,GHz band,
\begin{equation}
  |Z_{\mathrm{tr}}(i,j)|\;\approx\;Z_0\cdot\exp\!\bigl(-\alpha_x|x_i\!-\!x_j|-\alpha_y|y_i\!-\!y_j|\bigr),
  \label{eq:zdecay}
\end{equation}
with $\alpha_x,\alpha_y>0$ direction-specific decay constants (horizontal and vertical metal layers are typically asymmetric); a full derivation is in App.~\ref{app:pdn_physics}. In our Transformers, attention weights should mirror this exponential spatial decay, so we encode $\exp(-\alpha\cdot\dM)$ as a multiplicative bias on the linear-attention kernel; because the kernel separates along $x$ and $y$, it can be absorbed into $Q$ and $K$ separately, preserving linear complexity. All three of our scenarios fit this form. DPP and CircuitNet IR-drop are both built on PDN voltage distributions, so the derivation applies directly. Macro placement involves no electromagnetic coupling, but the same form is justified by routing locality: Rent's rule~\citep{landman1971rent} implies closer macros are more likely to share a net.

\subsubsection{Mathematical Construction of PSLA}
\label{sec:psla_math}

Linear attention~\citep{katharopoulos2020transformers} replaces the softmax-exponential kernel with a non-negative feature map $\phi:\mathbb R^d\to\mathbb R^d_+$ applied row-wise:
\begin{equation}
  \mathrm{LinAttn}(Q,K,V)=\frac{\phi(Q)[\phi(K)^\top V]}{\phi(Q)\phi(K)^\top\mathbf{1}}.
  \label{eq:linear_attention}
\end{equation}
Computing the $d\times d$ product $\phi(K)^\top V$ first avoids forming the $L\times L$ attention matrix, reducing total cost to $\mathcal{O}(Ld^2)$. The cost, however, is loss of softmax sharpness: weights diffuse toward uniformity and the model loses its local-structure inductive bias. PSLA introduces position-dependent multiplicative factors $D_Q,D_K\in\mathbb R^L$ that encode the distance decay as a structured bias on the kernel:
\begin{equation}
  \mathrm{PSLA}(Q,K,V)=\frac{(\phi(Q)\odot D_Q)[(\phi(K)\odot D_K)^\top V]}{(\phi(Q)\odot D_Q)(\phi(K)\odot D_K)^\top\mathbf{1}},\qquad
  \begin{aligned}
    D_Q[i]&=\exp(-\alpha_x x_i-\alpha_y y_i),\\
    D_K[j]&=\exp(+\alpha_x x_j+\alpha_y y_j),
  \end{aligned}
  \label{eq:psla}
\end{equation}
where $\odot$ is element-wise multiplication and $(x_i,y_i)\in[0,1]^2$ is the normalized chip coordinate. The effective pairwise weight decomposes as $a(i,j)\propto\phi(Q_i)^\top\phi(K_j)\cdot D_Q[i]D_K[j]$: the first factor captures \emph{what} is relevant, the second captures \emph{where} via a multiplicative soft mask.

Letting $\tilde Q_i=\phi(Q_i)\odot D_Q[i]$ and $\tilde K_j=\phi(K_j)\odot D_K[j]$, PSLA reduces to standard linear attention on $(\tilde Q,\tilde K,V)$, so the change-of-association in \eqref{eq:linear_attention} still applies and complexity remains $\mathcal{O}(Ld^2)$. Position information is encoded into $D_Q,D_K$ separately, without ever forming the $L\times L$ distance matrix, because $\exp(-\alpha\cdot\dM)$ separates along $x$ and $y$. Vanilla softmax attention, by contrast, has $\mathcal{O}(L^2 d)$ cost and quickly hits memory limits beyond $L\!\approx\!10{,}000$, while also relying entirely on content similarity $QK^\top$ and ignoring the distance prior. This rank-1 product is signed (directional) rather than the symmetric $\exp(-\alpha|x_j\!-\!x_i|)$, since absolute value admits no rank-1 factorization; an exact $\mathcal{O}(L)$ symmetric reconstruction via bidirectional prefix sums is in App.~\ref{app:bidirectional}.\footnote{Three mechanisms make the rank-1 form effective in practice: (i)~the denominator normalization acts like softmax and damps the unbounded direction; (ii)~multi-head attention learns complementary decay directions, with empirically observed coverage of both signs (App.~\ref{app:extended_ablations}); (iii)~causal masking in autoregressive decoders activates only one direction.}

\subsubsection{Learnable Decay Parameters}
\label{sec:psla_learnable}

The decay rates $\alpha_x,\alpha_y$ are not fixed hyperparameters but per-head, per-layer learnable scalars optimized by Adam alongside $Q,K,V$ projections. Crucially, $\alpha_x,\alpha_y$ are \emph{independent} parameters---unlike schemes that produce decay rates dynamically through a neural sub-module~\citep{mao2025lasad}. Two physical facts justify this. First, the form $\exp(-\alpha\cdot\dM)$ is fixed by the physical derivation \eqref{eq:zdecay}; only the rate $\alpha$ is learned. Second, $\alpha$ corresponds to the inverse PDN characteristic length $\lambda\!\equiv\!1/\alpha$, an instance-specific quantity~\citep{swaminathan2007power} that must be data-adaptive; across heads, small $\alpha$ captures global coupling, large $\alpha$ local refinement. For training stability we reparameterize via
$\alpha=\alpha_{\min}+(\alpha_{\max}-\alpha_{\min})\sigma(\alpha_{\mathrm{raw}})$
with $\sigma$ the sigmoid. Based on PDN impedance measurements we set $[\alpha_{\min},\alpha_{\max}]=[1.2,1.8]$ initialized to $1.5$. Each head adds two scalars ($48$ total in our 8-head 3-layer model), and the physically-motivated narrow range itself acts as effective regularization (\S\ref{sec:ablation}). PSLA also includes a data-dependent gate based on gated linear attention~\citep{yang2024gla}; we initialize it near-closed so the distance-decay prior dominates early training and yields to data as it proceeds (App.~\ref{app:feature_map}).

Table~\ref{tab:complexity} compares PSLA's complexity to vanilla and linear attention; the only overhead beyond linear attention is the linear-time $D_Q,D_K$ multiplication. ALiBi~\citep{press2022train} adds a fixed-slope linear distance penalty to causal scores; RoPE~\citep{su2024roformer} encodes absolute positions via rotation matrices; both target 1D token order on softmax attention without reducing complexity. PSLA extends the idea to 2D Manhattan-distance space defined by physical coordinates, fuses it multiplicatively into the linear-attention kernel, and makes decay rates per-head learnable.

\begin{table}[h]
\centering
\caption{Attention-mechanism complexity. $L$ is sequence length, $d$ feature dimension, $N$ number of layers.}
\label{tab:complexity}
\small
\begin{tabular}{@{}lccccc@{}}
\toprule
& \multicolumn{2}{c}{\textbf{Train (per encoder layer)}} & \multicolumn{2}{c}{\textbf{Inference}} & \textbf{Memory} \\
\cmidrule(lr){2-3}\cmidrule(lr){4-5}
\textbf{Mechanism} & Time & Space & Encoder total & Decoder/step & Attention \\
\midrule
Softmax & $\mathcal{O}(L^2 d)$ & $\mathcal{O}(L^2)$ & $\mathcal{O}(NL^2 d)$ & $\mathcal{O}(Ld)$ & $\mathcal{O}(L^2)$ \\
Linear  & $\mathcal{O}(Ld^2)$  & $\mathcal{O}(Ld)$  & $\mathcal{O}(NLd^2)$  & $\mathcal{O}(Ld)$ & $\mathcal{O}(Ld)$ \\
\textbf{PSLA} & $\mathcal{O}(Ld^2)$ & $\mathcal{O}(Ld)$ & $\mathcal{O}(NLd^2)$ & $\mathcal{O}(Ld)$ & $\mathcal{O}(Ld)$ \\
\bottomrule
\end{tabular}
\end{table}

\subsection{PBRS: Potential-Based Reward Shaping}
\label{sec:pbrs}

RL on EDA tasks suffers extreme reward sparsity: in DPP the simulator returns a non-zero reward only after all $K\!=\!25$ capacitors are placed via Kron reduction~\citep{dorfler2013kron}; intermediate steps yield no signal (App.~\ref{app:dpp_reward}). PBRS~\citep{ng1999policy} replaces the original $R(s,a,s')$ with $R'(s,a,s')=R(s,a,s')+\gamma\Phi(s')-\Phi(s)$, where $\Phi:\mathcal{S}\!\to\!\mathbb{R}$ scores physical plausibility. The policy-invariance theorem guarantees that for any $\Phi$ the optimal policy is unchanged: summed over a trajectory the shaping term telescopes to $\gamma^T\Phi(s_T)-\Phi(s_0)$, depending only on endpoints (App.~\ref{app:pbrs_invariance}). Any physically plausible $\Phi$ thus provides dense intermediate signal without bias.

For DPP, the potential $\Phi_{\mathrm{DPP}}$ combines two physics-grounded terms based on the same $\exp(-\alpha\cdot\dM)$ kernel:
\begin{equation}
  \Phi_{\mathrm{DPP}}(s)=\sum_{i\in\mathcal{P}}\exp(-\alpha\,\dM(i,p))\;-\;\lambda\sum_{\substack{i,j\in\mathcal{P}\\i<j}}\exp(-\alpha\,\dM(i,j)),
  \label{eq:phi_dpp}
\end{equation}
where $\mathcal{P}$ is the placed set and $p$ the probe port. The first term encourages capacitors near the probe (local impedance falls with distance); the second penalizes clustering (closely co-located capacitors share the same discharge path with diminishing returns). $\Phi_{\mathrm{DPP}}$ uses the same kernel as the PSLA bias, doubly encoding one physical regularity. For macro placement, PBRS engages only during online RL fine-tuning (REINFORCE/GRPO~\citep{shao2024deepseekmath}), since offline Decision-Transformer pretraining~\citep{chen2021decision} does not interact step-wise with the environment. We use a connectivity-based potential $\Phi_{\mathrm{chip,conn}}(s)=\sum_{e\in\mathcal{E}}w_e\sum_{i,j\in e\cap\mathcal{P}}\exp(-\alpha\,\dM(i,j))$ that mirrors the HPWL contribution of placed pairs (App.~\ref{app:potential_equivalence}).

While policy invariance guarantees the optimum is unchanged, in practice a too-strong shaping term can dominate the original reward and slow learning. We multiply the shaping by a time-dependent weight $\beta(t)$ that decays cosine-style from $\beta_{\mathrm{init}}$ toward a small $\beta_{\min}$ (App.~\ref{app:beta_schedule}); conceptually a curriculum that lets the physical prior dominate early and yield to the simulator signal as it sharpens.

\subsection{Modular Integration}
\label{sec:unified_block}

PSLA and PBRS plug into existing EDA models without architectural surgery. PSLA replaces the attention layer while keeping embeddings, feed-forwards, and decoder structure intact, requiring only the physical coordinates $(x_i,y_i)$ of each token to compute $D_Q,D_K$. We instantiate PSLA in three architectures: DevFormer's autoregressive decoder (\S\ref{sec:exp_dpp}), ChiPFormer's Decision Transformer (\S\ref{sec:exp_chipformer}), and a UNet bottleneck (\S\ref{sec:exp_circuitnet}). PBRS modifies only the reward; the potential $\Phi$ supports incremental computation at $\mathcal{O}(K)$ per step, and at $\gamma\!=\!1$ collapses to $\beta[\Phi(s_T)-\Phi(s_0)]$. Policy invariance holds for any MDP algorithm; we verify on REINFORCE (DPP) and GRPO (ChiPFormer).

\section{Related Work}
\label{sec:related}

Three generations have driven learning-based macro placement: \citet{mirhoseini2021graph} and AlphaChip~\citep{mirhoseini2024alphachip} first applied GNN+RL with hypergraph netlists; MaskPlace~\citep{lai2022maskplace} switched to a 2D pixel canvas with dense rewards; ChiPFormer~\citep{lai2023chipformer} reformulated placement as offline RL via a Decision Transformer. Recent extensions push along target functions (LaMPlace~\citep{li2025lamplace} drives toward post-routing timing), data sources (diffusion-based~\citep{lee2025diffusion} for zero-shot transfer), and policy positioning (RL regulators~\citep{xue2024maskregulate} adjust existing layouts). For DPP, DevFormer~\citep{kim2023devformer} uses symmetry plus learned relative position encoding tied to fixed grid sizes, without a Manhattan-distance prior, limiting cross-scale transfer. For IR-drop prediction, PowerNet~\citep{xie2020powernet}, IREDGe~\citep{chhabria2021iredge}, and PDNNet~\citep{zhao2024pdnnet} progressively add CNN, image-translation, and dual-branch GNN+CNN designs. PDNNet also injects PDN structure but via an auxiliary GNN; PSLA folds the same prior into the attention kernel as a multiplicative bias---no extra module, linear complexity.

For efficient attention, Linformer~\citep{wang2020linformer} compresses keys/values via low-rank projection; FAVOR+~\citep{choromanski2021rethinking} uses random features; linear attention~\citep{katharopoulos2020transformers} replaces softmax with kernel decomposition for exact $\mathcal{O}(Ld^2)$ cost (PSLA's base) but offers no spatial prior; CosFormer~\citep{qin2022cosformer} adds a cosine local re-weighting on 1D token order; all target 1D sequences. Gated linear attention~\citep{yang2024gla} introduces data-dependent gates for content-based decay; PSLA reuses the gate but supplies a physics-driven, not data-driven, decay. Concurrent and independent of our work, LASAD~\citep{mao2025lasad} introduces a learnable 2D spatial decay in linear attention for image generation, with data-driven parameterization in a different domain.

\section{Experiments}
\label{sec:experiments}

We validate PhysEDA on three chip-design scenarios spanning problem types and architectures: (1) decoupling-capacitor placement (DPP), a sequential-decision task with an autoregressive decoder, trained by imitation learning and REINFORCE online RL; (2) macro placement, a sequential-decision task on continuous coordinates, evaluated using the ChiPFormer~\citep{lai2023chipformer} Decision-Transformer architecture with offline pretraining followed by online RL fine-tuning; (3) IR-drop prediction, a pixel-level regression task on the CircuitNet~\citep{chai2022circuitnet} benchmark with a UNet architecture. All experiments use the same PSLA implementation with $\alpha\in[1.2,1.8]$ initialized to $1.5$; full hyperparameters are in App.~\ref{app:hyperparams}.

\subsection{Scenario 1: Decoupling Capacitor Placement}
\label{sec:exp_dpp}

DPP places a fixed number of decoupling capacitors on a chip grid to minimize the integrated transfer impedance from each probe port over the operating band, reported as the DPP cost (negative-valued, more negative is better) following \citet{kim2023devformer}. Our setting differs from DevFormer in three ways: we extend evaluation from $10\!\times\!10$ ($L\!=\!100$, 25 capacitors) to $25\!\times\!25$ ($L\!=\!625$, 101 capacitors); we add REINFORCE online RL on top of imitation learning to isolate the architectural contribution of PSLA from the training-stage contribution of PBRS; and unlike DevFormer's setup which lacks a held-out validation set, we adopt a strict $2{,}000/100/200$ train/val/test split for both supervised and RL experiments on a $2{,}300$-instance dataset generated by DevFormer's physical formula. Full baseline list, dataset construction, split protocol, and $N_{\mathrm{aug}}\!=\!4$ test-time augmentation are in App.~\ref{app:exp_dpp}.

\begin{table}[h]
  \centering
  \caption{DPP imitation learning on $25\times 25$. Lower (more negative) is better. PSLA is the only linear-complexity method to beat the quadratic baseline.}
  \label{tab:dpp_il}
  \small
  \begin{tabular}{lcc}
    \toprule
    \textbf{Method} & \textbf{Complexity} & \textbf{Score ($\downarrow$)} \\
    \midrule
    \textbf{PSLA (ours)} & $\mathcal{O}(Ld^2)$ & $\mathbf{-16.76}$ \\
    DevFormer~\citep{kim2023devformer} & $\mathcal{O}(L^2 d)$ & $-15.88$ \\
    Plain GLA~\citep{yang2024gla} & $\mathcal{O}(Ld^2)$ & $-15.81$ \\
    FAVOR+~\citep{choromanski2021rethinking} & $\mathcal{O}(Ld^2)$ & $-15.79$ \\
    Simple Linear~\citep{katharopoulos2020transformers} & $\mathcal{O}(Ld^2)$ & $-15.67$ \\
    CosFormer~\citep{qin2022cosformer} & $\mathcal{O}(Ld^2)$ & $-15.48$ \\
    \midrule
    \multicolumn{2}{l}{\textit{PSLA vs.\ DevFormer}} & \textit{$+5.5\%$} \\
    \bottomrule
  \end{tabular}
\end{table}

Table~\ref{tab:dpp_il} shows PSLA scores $-16.76$ on $25\!\times\!25$, beating quadratic-complexity DevFormer ($-15.88$) by $5.5\%$ and the four other linear-attention variants by $6.0\%\!-\!8.3\%$. Notably, all the other linear-attention variants fall \emph{below} DevFormer, confirming that reducing complexity without a spatial prior is not enough.

\begin{figure}[h]
  \centering
  \begin{minipage}[t]{0.48\linewidth}
    \centering
    \includegraphics[width=\linewidth]{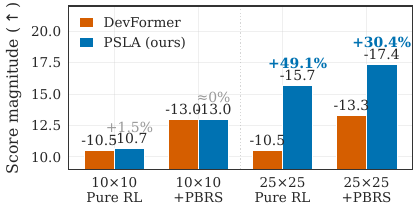}
    \caption{DPP RL with PBRS. Going from $L\!=\!100$ ($10\!\times\!10$) to $L\!=\!625$ ($25\!\times\!25$), PSLA's gain jumps from $1.5\%$ to $49.1\%$.}
    \label{fig:dpp_rl}
  \end{minipage}\hfill
  \begin{minipage}[t]{0.48\linewidth}
    \centering
    \includegraphics[width=\linewidth]{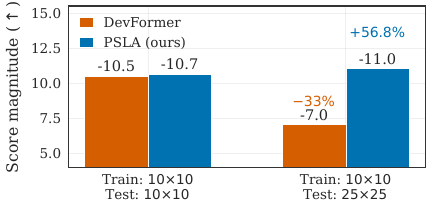}
    \caption{Zero-shot cross-scale transfer ($10\!\times\!10\!\to\!25\!\times\!25$). DevFormer degrades by $33\%$, PSLA by $3.6\%$, a $56.8\%$ relative gain.}
    \label{fig:scale_gen}
  \end{minipage}
\end{figure}

Table~\ref{tab:dpp_rl} and Figure~\ref{fig:dpp_rl} contrast REINFORCE results at two scales---the cleanest evidence that PSLA's gain monotonically tracks data insufficiency. At $10\!\times\!10$ ($L\!=\!100$) the search space is small enough that REINFORCE explores adequately; PSLA's gain is only $1.5\%$, and after PBRS the two are tied at $-13.01$ (the prior is essentially redundant). At $25\!\times\!25$ ($L\!=\!625$) the sequence is $6\times$ longer and exploration becomes sparse; PSLA delivers a $49.1\%$ gain, and even after both add PBRS, PSLA retains a $30.4\%$ advantage. The two scales give $1.5\%$ vs.\ $49.1\%$, a textbook illustration of the data-insufficiency principle.

\begin{table}[h]
  \centering
  \caption{DPP REINFORCE results, $10\times 10$ and $25\times 25$. PBRS uses $\Phi_{\mathrm{DPP}}$ from \eqref{eq:phi_dpp} with cosine $\beta$ schedule.}
  \label{tab:dpp_rl}
  \small
  \begin{tabular}{lcccc}
    \toprule
    & \multicolumn{2}{c}{$10\times 10$} & \multicolumn{2}{c}{$25\times 25$} \\
    \cmidrule(lr){2-3}\cmidrule(lr){4-5}
    \textbf{Configuration} & DevFormer & PSLA & DevFormer & PSLA \\
    \midrule
    No shaping (vanilla RL) & $-10.50$ & $-10.66$ & $-10.53$ & $\mathbf{-15.70}$ \\
    + PBRS                  & $-13.01$ & $-13.01$ & $-13.34$ & $\mathbf{-17.40}$ \\
    \midrule
    PSLA gain (vanilla RL)  & \multicolumn{2}{c}{$+1.5\%$} & \multicolumn{2}{c}{$+49.1\%$} \\
    PSLA gain (+PBRS)       & \multicolumn{2}{c}{$\approx 0\%$} & \multicolumn{2}{c}{$+30.4\%$} \\
    \bottomrule
  \end{tabular}
\end{table}

We also test cross-scale zero-shot transfer using vanilla-RL models without PBRS, so that the comparison isolates PSLA's architectural contribution rather than any training-stage effect: we train on $10\!\times\!10$ and evaluate directly on $25\!\times\!25$. DevFormer drops from $-10.50$ to $-7.04$, a $33\%$ absolute decline, while PSLA goes from $-10.66$ to $-11.04$, marginally better at the larger scale, a $56.8\%$ relative improvement over DevFormer (Figure~\ref{fig:scale_gen}). The reason is that the Manhattan-distance decay encoded by PSLA is scale-invariant by construction: $\exp(-\alpha\cdot\dM)$ depends only on the absolute coordinate differences, and the learned $\alpha$ corresponds to the inverse PDN characteristic length $\lambda\!\equiv\!1/\alpha$, set by per-unit-length resistance and inductance of the metal stack and the operating frequency band, not by grid resolution. Once this rate is fitted at $10\!\times\!10$ it remains correct at $25\!\times\!25$, so PSLA does not need to relearn spatial structure; DevFormer encodes spatial relations only through learned relative position embeddings whose semantics shift with grid size, so its representation does not transfer.

\subsection{Scenario 2: Macro Placement (ChiPFormer)}
\label{sec:exp_chipformer}

Macro placement places a fixed set of large IP modules onto a chip canvas at continuous coordinates, optimizing the half-perimeter wirelength (HPWL): the sum over all nets of the half-perimeter of the bounding box of the net's terminals, a standard wirelength proxy in physical placement. We integrate PSLA into the ChiPFormer architecture~\citep{lai2023chipformer} by replacing only its GPT-style attention layer, and evaluate on the ISPD 2005 \texttt{adaptec1} benchmark with 452 macros, using ChiPFormer's public offline expert trajectories for pretraining and online interaction for RL fine-tuning. Baselines include GPT DT for pretraining and GPT REINFORCE, GPT GRPO~\citep{shao2024deepseekmath}, and GPT with PBRS as a logit bias for RL fine-tuning; full setup is in App.~\ref{app:exp_chipformer}. Lower HPWL is better.

Table~\ref{tab:chipformer_pretrain} shows that at 300 epochs, replacing GPT's vanilla attention with PSLA reduces HPWL by $8.1\%\!-\!12.0\%$: the connectivity-aware variant reaches $801{,}588$ while GPT plateaus at $910{,}967$.\footnote{At 1{,}000 epochs both converge to $801{,}588$, exactly as our unified framework predicts: with sufficient training, the data-driven model eventually learns the spatial structure on its own and the prior provides no extra benefit. In practical settings with budget around 300--500 epochs, PSLA's $12\%$ improvement amounts to roughly a $3\times$ training-efficiency gain.}

\begin{table}[h]
  \centering
  \caption{ChiPFormer Decision-Transformer pretraining on \texttt{adaptec1}, 300 epochs.}
  \label{tab:chipformer_pretrain}
  \small
  \begin{tabular}{lcc}
    \toprule
    \textbf{Method} & \textbf{HPWL ($\downarrow$)} & \textbf{vs.\ GPT DT} \\
    \midrule
    \textbf{Connectivity PSLA DT (ours)} & $\mathbf{801{,}588}$ & $+12.0\%$ \\
    PSLA DT (ours)                       & $837{,}584$            & $+8.1\%$ \\
    GPT DT (baseline)                    & $910{,}967$            & --- \\
    \bottomrule
  \end{tabular}
\end{table}

Table~\ref{tab:chipformer_rl} reports RL fine-tuning, where \emph{GPT + PBRS logit bias} achieves the best HPWL at $693{,}999$, a $5.4\%$ improvement over GPT REINFORCE. Notably, PSLA + GRPO scores $762{,}974$, \emph{worse} than GPT + GRPO. The reason is constraint type: PSLA imposes a hard architectural constraint that helps DT pretraining converge but limits RL exploration, while PBRS's logit bias is a soft constraint that decays through $\beta$ scheduling and acts at action probabilities. The two integration paths thus complement: PSLA helps pretraining, PBRS helps RL fine-tuning.

\begin{table}[h]
  \centering
  \caption{ChiPFormer RL fine-tuning on \texttt{adaptec1}. All start from the GPT DT checkpoint.}
  \label{tab:chipformer_rl}
  \small
  \begin{tabular}{lcc}
    \toprule
    \textbf{Method} & \textbf{Best HPWL ($\downarrow$)} & \textbf{vs.\ GPT REINFORCE} \\
    \midrule
    \textbf{GPT + PBRS logit bias (ours)} & $\mathbf{693{,}999}$ & $\mathbf{+5.4\%}$ \\
    GPT REINFORCE                         & $733{,}794$            & --- \\
    GPT GRPO                              & $748{,}241$            & $-2.0\%$ \\
    GPT GRPO + PBRS                       & $751{,}804$            & $-2.5\%$ \\
    PSLA + GRPO                           & $762{,}974$            & $-4.0\%$ \\
    \bottomrule
  \end{tabular}
\end{table}

\subsection{Scenario 3: IR-Drop Prediction (CircuitNet)}
\label{sec:exp_circuitnet}

IR-drop prediction takes a chip's power-density map as input and predicts the per-pixel voltage drop on the PDN; excessive drop causes timing failures, so accurate prediction is critical for verification. We use a standard UNet image-to-image translation baseline and replace only its convolutional bottleneck with a single PSLA attention block (PSLA-UNet), keeping all other layers identical. Evaluation uses CircuitNet 1.0~\citep{chai2022circuitnet} (native resolution $459\!\times\!456$) under three settings, each with five random seeds: in-distribution Vortex; cross-design across three directions; and cross-architecture across three directions (App.~\ref{app:exp_circuitnet}). The metric is mean Pearson correlation between predicted and ground-truth IR-drop maps with the per-seed win rate against the baseline; higher is better.

Table~\ref{tab:circuitnet_all} validates the framework's prediction by contrasting data-rich and data-poor settings. In the in-distribution setting, PSLA is slightly behind by $-1.9\%$ (paired $t\!=\!-5.17$ across 5 seeds, $p\!<\!0.005$), a small but statistically significant deficit consistent with the framework: when training data is sufficient for the baseline to learn the spatial structure on its own, the additional physical constraint hurts capacity. Under distribution shift the picture flips. Cross-design improves $+5.3\%$ with a 12-of-15 win rate; cross-architecture improves $+5.4\%$ with an 11-of-15 win rate; per-direction breakdowns are in App.~\ref{app:extended_ablations}.

\begin{table}[h]
  \centering
  \caption{CircuitNet IR-drop prediction. Pearson correlation (higher is better). PSLA loses slightly in-distribution (statistically significant, $p\!<\!0.005$$^{*}$) and consistently wins under distribution shift. ${}^{*}$paired $t$-test on 5 seeds; full per-seed values in App.~\ref{app:extended_ablations}.}
  \label{tab:circuitnet_all}
  \small
  \begin{tabular}{lcccc}
    \toprule
    \textbf{Setting} & \textbf{Baseline} & \textbf{PSLA} & \textbf{$\Delta$} & \textbf{Win rate} \\
    \midrule
    In-distribution Vortex (5 seeds)        & $0.834\!\pm\!0.036$ & $0.818\!\pm\!0.040$ & $-1.9\%^{*}$ & --- \\
    Cross-design (3 directions, 5 seeds)    & $0.458$ & $\mathbf{0.482}$ & $\mathbf{+5.3\%}$ & \textbf{12/15} \\
    Cross-architecture (3 directions, 5 seeds) & $0.472$ & $\mathbf{0.498}$ & $\mathbf{+5.4\%}$ & \textbf{11/15} \\
    \bottomrule
  \end{tabular}
\end{table}

\subsection{Efficiency Analysis}
\label{sec:efficiency}

PSLA's practical efficiency depends on $L$: the larger $L$, the more linear complexity dominates the quadratic baseline. Table~\ref{tab:efficiency} summarizes RTX 4090 single-batch measurements across the three scenarios; the full latency-and-memory curve versus $L$ is in App.~\ref{app:extended_ablations}. For the DPP encoder the crossover is at $50\!\times\!50$ with $L\!=\!2{,}500$, and at $100\!\times\!100$ PSLA achieves a $14\times$ speedup with $98.5\%$ memory savings while using only 65\,MB. At $150\!\times\!150$ softmax requires roughly 30\,GB and exceeds most GPUs, while PSLA still runs at $32.5\times$ speedup. Full-model end-to-end speedup at $100\!\times\!100$ is $3.2\times$; the shared decoder dilutes the encoder gain. For ChiPFormer, attention-layer memory drops by $94\%$ ($1.2$\,GB$\to 68$\,MB) with break-even latency at $T\!\approx\!512$. For CircuitNet, PSLA adds a small overhead in this attention-light regime. The three scenarios thus cover three regimes: DPP is far above the crossover, ChiPFormer is around it, and CircuitNet is below it.

\begin{table}[h]
  \centering
  \caption{Inference efficiency summary; RTX 4090, single-batch.}
  \label{tab:efficiency}
  \small
  \begin{tabular}{@{}llcc@{}}
    \toprule
    \textbf{Scenario} & \textbf{Operating point} & \textbf{Speed} & \textbf{Memory} \\
    \midrule
    DPP encoder    & $100\!\times\!100$, $L\!=\!10{,}000$ & $\mathbf{14\times}$ & $-98.5\%$ \\
    DPP full       & $100\!\times\!100$, $L\!=\!10{,}000$ & $3.2\times$ & $-75\%$ \\
    ChiPFormer     & $T\approx 512$ & $1.08\times$ (48 vs.\ 52\,ms) & $-94\%$ ($1.2$\,GB$\to$68\,MB) \\
    CircuitNet     & $N=256$ bottleneck & $+2.4$\,ms overhead & $+2.4$\,MB \\
    \bottomrule
  \end{tabular}
\end{table}

\subsection{Unified Ablation Across Three Scenarios}
\label{sec:ablation}

Table~\ref{tab:unified_ablation} reports PSLA and PBRS contributions across the three scenarios. For each setting we report four configurations whenever applicable: a vanilla baseline that uses neither PSLA nor PBRS, a PSLA-only variant that adds the architectural prior alone, a PBRS-only variant that adds the training-stage shaping alone, and the full PSLA+PBRS combination. The last column reports the best \% improvement over vanilla: positive = improvement, negative = degradation.

\begin{table}[h]
  \centering
  \caption{Unified ablation of PSLA and PBRS across DPP, ChiPFormer, and CircuitNet. PSLA adds the architectural distance prior; PBRS adds the training-stage potential shaping. Cells marked n/a denote supervised settings to which PBRS does not apply. Best configuration per row in bold; ${}^{\dagger}$ indicates that PSLA's hard architectural constraint suppresses exploration during RL fine-tuning, see Section~\ref{sec:exp_chipformer}. Two finer-grained DPP ablations supporting these aggregate numbers (component decomposition and $\alpha$-range sensitivity) are deferred to App.~\ref{app:extended_ablations}.}
  \label{tab:unified_ablation}
  \footnotesize
  \begin{tabular}{@{}llccccc@{}}
    \toprule
    \textbf{Scenario} & \textbf{Type} & \textbf{Vanilla} & \textbf{+\,PSLA} & \textbf{+\,PBRS} & \textbf{+\,PSLA\,+\,PBRS} & \textbf{Best $\Delta$} \\
    \midrule
    DPP IL ($25\!\times\!25$)        & supervised        & $-15.88$    & $\mathbf{-16.76}$            & n/a              & n/a               & $+5.5\%$  \\
    DPP RL ($10\!\times\!10$)        & online RL         & $-10.50$    & $-10.66$                     & $\mathbf{-13.01}$& $-13.01$          & $+23.9\%$ \\
    DPP RL ($25\!\times\!25$)        & online RL         & $-10.53$    & $-15.70$                     & $-13.34$         & $\mathbf{-17.40}$ & $+65.2\%$ \\
    DPP transfer ($10\!\to\!25$)     & zero-shot         & $-7.04$     & $\mathbf{-11.04}$            & n/a              & n/a               & $+56.8\%$ \\
    \midrule
    ChiPFormer DT                    & supervised        & $910{,}967$ & $\mathbf{801{,}588}$         & n/a              & n/a               & $+12.0\%$ \\
    ChiPFormer RL fine-tune          & online RL         & $733{,}794$ & $762{,}974^{\dagger}$        & $\mathbf{693{,}999}$ & $763{,}009^{\dagger}$ & $+5.4\%$  \\
    \midrule
    CircuitNet cross-design          & regression        & $0.458$     & $\mathbf{0.482}$             & n/a              & n/a               & $+5.3\%$  \\
    CircuitNet cross-architecture    & regression        & $0.472$     & $\mathbf{0.498}$             & n/a              & n/a               & $+5.4\%$  \\
    CircuitNet in-distribution       & regression        & $\mathbf{0.834}$ & $0.818$                 & n/a              & n/a               & $-1.9\%$  \\
    \bottomrule
  \end{tabular}
\end{table}

Read along the rows, the table reproduces the unified data-insufficiency principle: $|\Delta|$ rises monotonically with data insufficiency for spatial structure, from $-1.9\%$ (CircuitNet in-distribution, prior redundant), through $+5\%$ to $+12\%$ (CircuitNet cross-domain, DPP supervised, ChiPFormer DT pretraining), to $+56.8\%$ (cross-scale transfer) and $+65.2\%$ (DPP RL, larger grid, prior carries full load). Read down the columns, the +\,PSLA column captures the architectural contribution and dominates in supervised and zero-shot transfer settings, while the +\,PBRS column captures the training-stage contribution and dominates in RL settings; the two components are additive on DPP $25\!\times\!25$ but conflict on ChiPFormer RL (PSLA's hard constraint vs PBRS's soft schedule); this motivates the two-component design.

\subsection{Unified Data-Insufficiency Principle}
\label{sec:unified_analysis}
\label{sec:interpretability}

\begin{figure}[h]
  \centering
  \includegraphics[width=\linewidth]{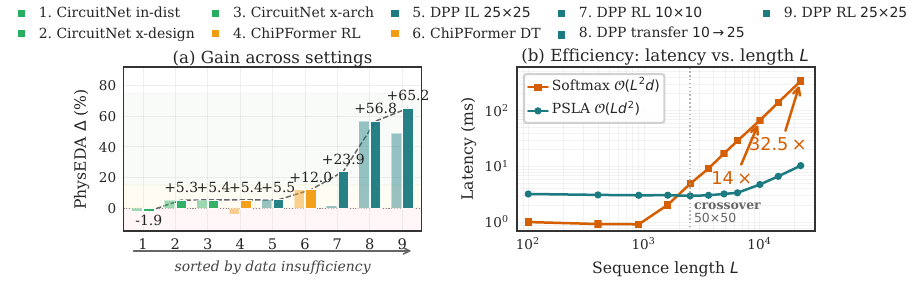}
  \caption{PhysEDA $\Delta$ across 9 settings (a) and inference latency vs.\ $L$ (b). In (a), light bars are $+$PSLA only and dark bars the best (PSLA/PBRS) combination, ranging from $-1.9\%$ to $+65.2\%$. In (b), RTX 4090 batch-1; PSLA's $\mathcal{O}(Ld^2)$ stays flat while softmax $\mathcal{O}(L^2 d)$ rises sharply, with crossover at $L\!=\!2{,}500$.}
  \label{fig:unified_narrative}
\end{figure}

Figure~\ref{fig:unified_narrative}\,(a) visualizes the same monotonic pattern across all three scenarios, from $-1.9\%$ (data-rich, prior redundant) to $+65.2\%$ (DPP RL with the larger grid). Figure~\ref{fig:unified_narrative}\,(b) reports the corresponding efficiency: PSLA's $\mathcal{O}(Ld^2)$ scaling crosses the softmax baseline at $L\!=\!2{,}500$ ($50\!\times\!50$) and reaches $14\times$ speedup at $L\!=\!10{,}000$ and $32.5\times$ at $L\!=\!22{,}500$, where softmax exceeds 30\,GB and runs out of memory. The same physical prior thus delivers two compounding benefits---accuracy where data is scarce, and tractable computation where the design is large.

\section{Conclusion}
\label{sec:conclusion}

PhysEDA integrates a single Manhattan-distance physical prior at both architecture (PSLA, an $\mathcal{O}(Ld^2)$ multiplicative bias on linear attention) and training (PBRS, a potential from the same kernel that preserves policy invariance). Across three EDA scenarios, the gain monotonically tracks data insufficiency for spatial structure, ranging from $-1.9\%$ in-distribution to $+56.8\%$ on zero-shot cross-scale transfer; PSLA dominates supervised settings, PBRS dominates RL exploration. Two limitations remain: the rank-1 factorization realizes directional rather than symmetric Manhattan decay (App.~\ref{app:bidirectional} gives an exact $\mathcal{O}(L)$ reconstruction), and PBRS applies only to online RL. Promising follow-ups include the bidirectional decomposition, broader physical potentials such as parasitic capacitance and signal-integrity margins, and extending the framework to clock distribution and other EDA tasks.

\begin{ack}
The author thanks Xin Shen and Lingfeng Niu (Chinese Academy of Sciences) for
guidance on research direction, technical discussions on physical motivation
and mathematical construction, and detailed feedback on the experimental
design throughout this work. The author also thanks Qingna Li (Beijing
Institute of Technology) for supervision and review of the manuscript.
The author declares no competing interests.
\end{ack}

\bibliographystyle{plainnat}
\bibliography{references}

\appendix
%
%

\section{PDN Physics: Manhattan Distance Decay Derivation}
\label{app:pdn_physics}


\paragraph{Setup: PDN as a 2D transmission-line mesh.}
Following \citet{swaminathan2007power}, we model the on-chip power-delivery network (PDN) as a two-dimensional mesh of transmission-line segments along the orthogonal metal layers. Each segment carries per-unit-length resistance $R$, inductance $L$, and capacitance $C$. At frequency $f$ with $\omega = 2\pi f$, the segment presents a complex propagation constant $\gamma(f) = \sqrt{(R+j\omega L)\,j\omega C} = \kappa(f) + j\beta(f)$. The real part $\kappa(f) > 0$ is the per-segment attenuation: a wave entering with unit amplitude exits with amplitude $e^{-\kappa(f)}$.

\paragraph{Exponential decay along the path.}
Consider the transfer impedance $Z_{\mathrm{tr}}(i,j,f)$ — the voltage response at node $i$ to unit current injection at node $j$. For a path of $n$ segments between $j$ and $i$, the cumulative attenuation gives
\begin{equation}
  |Z_{\mathrm{tr}}(i,j,f)| \;\propto\; \exp\!\bigl(-\kappa(f)\,n\bigr).
  \label{eq:zexp_app}
\end{equation}
This is the multi-segment generalization of the elementary lossy-transmission-line result. On standard PDN test cases the transmission-line model agrees numerically with the cavity-resonance formulation \citep{kim2010chip} to within a few percent over the 100\,MHz--2\,GHz band relevant for decoupling \citep{kim2001transmission}, so either model can serve as the physical basis.

\paragraph{Path length under orthogonal routing.}
Modern VLSI uses orthogonal routing: alternating metal layers carry horizontal and vertical wires only, both for crosstalk reduction and for via-count minimization \citep{schulte2012design,koh2000manhattan_routing}. Current from $j$ to $i$ therefore cannot follow a Euclidean diagonal but must traverse a staircase of horizontal and vertical segments, with total path length equal to the Manhattan distance:
\begin{equation}
  n \;=\; \dM(i,j) \;\equiv\; |x_i - x_j| + |y_i - y_j|.
  \label{eq:nmanhattan}
\end{equation}

\paragraph{Combining: Manhattan exponential decay.}
Substituting \eqref{eq:nmanhattan} into \eqref{eq:zexp_app}, and noting that horizontal and vertical metal layers are typically asymmetric (different per-unit-length $R$, $L$ due to wire width and pitch differences), we split the per-segment attenuation into direction-specific constants $\alpha_x \equiv \kappa_h(f)$ and $\alpha_y \equiv \kappa_v(f)$. Because the exponential of a sum factors as a product, this is exact:
\begin{equation}
  |Z_{\mathrm{tr}}(i,j,f)| \;\approx\; Z_0(f) \cdot \exp\!\bigl(-\alpha_x|x_i-x_j| - \alpha_y|y_i-y_j|\bigr),
  \label{eq:zdecay_app}
\end{equation}
recovering Eq.~\ref{eq:zdecay} of the main paper. Here $Z_0(f)$ is the zero-distance reference impedance and $\alpha_x, \alpha_y > 0$. The decomposition along $x$ and $y$ is the property that makes PSLA's rank-1 absorption into $D_Q, D_K$ possible while preserving linear complexity.

\paragraph{Empirical support.}
\citet{fan2000quantifying} provide direct experimental support: full-wave electromagnetic simulation on representative PDN test vehicles showed that impedance reduction at the probe port drops monotonically and approximately exponentially with capacitor distance over the 100\,MHz--2\,GHz band, vanishing at centimetre-scale separations.

\paragraph{Application beyond DPP.}
Two of our three scenarios share Eq.~\ref{eq:zdecay_app} directly. CircuitNet IR-drop prediction is governed by the same PDN voltage propagation: the network's input is per-pixel current density and the output is the per-pixel voltage drop satisfying the same transmission-line equations on the same orthogonal mesh \citep{zhao2024pdnnet}, so $\exp(-\alpha\dM)$ applies pixel-to-pixel without modification. Macro placement does not involve electromagnetic coupling, but the same form is justified by routing locality: Rent's rule \citep{landman1971rent} relates block size and IO count, with the corollary that the probability that two macros share a net decreases monotonically with their L1 separation; closer macros are statistically more likely to be connected by short wires. The functional form remains $\exp(-\alpha\dM)$, but its physical basis shifts from electromagnetic propagation to net-sharing statistics.

\section{Bidirectional Prefix-Sum Decomposition}
\label{app:bidirectional}


The rank-1 factorization $D_Q[i] \cdot D_K[j] = \exp\bigl(\alpha(x_j - x_i)\bigr)$ used in \S\ref{sec:psla_math} of the main paper realizes a directional, signed exponential rather than the symmetric Manhattan decay $\exp(-\alpha|x_i - x_j|)$. The reason is geometric: $|x_i - x_j|$ is not a bilinear function of $(x_i, x_j)$, so it admits no rank-1 factorization $f(x_i)\,g(x_j)$. The cusp at $x_i = x_j$ separates the regimes $x_i \geq x_j$ versus $x_i < x_j$, and any single rank-1 product is smooth across this point and so cannot reproduce the jump in derivative.

\paragraph{Decomposition.}
The symmetric kernel can nonetheless be reconstructed in $\mathcal{O}(L)$ by splitting the sum over key positions into two causal halves on either side of the query:
\begin{equation}
  \sum_{j} \exp\!\bigl(-\alpha|x_i - x_j|\bigr)\,\phi(K_j)^\top V_j
  = \underbrace{\sum_{j \leq i} \exp\!\bigl(-\alpha(x_i - x_j)\bigr)\,\phi(K_j)^\top V_j}_{S^{+}_i \text{ (forward pass)}}
  + \underbrace{\sum_{j > i} \exp\!\bigl(-\alpha(x_j - x_i)\bigr)\,\phi(K_j)^\top V_j}_{S^{-}_i \text{ (backward pass)}}.
  \label{eq:bidir_decomp}
\end{equation}
Each pass is a standard causal linear attention with one-directional exponential decay, and computes via a simple recurrence:
\begin{align}
  S^{+}_i &= \exp\!\bigl(-\alpha(x_i - x_{i-1})\bigr)\,S^{+}_{i-1} + \phi(K_i)^\top V_i,\\
  S^{-}_i &= \exp\!\bigl(-\alpha(x_{i+1} - x_i)\bigr)\,S^{-}_{i+1} + \phi(K_{i+1})^\top V_{i+1},
\end{align}
each with $\mathcal{O}(d^2)$ work per step and $\mathcal{O}(L d^2)$ total. The two passes can be run in parallel — forward sweep on $\phi(K)$ in original order, backward sweep on $\phi(K)$ in reversed order — so the wall-clock cost is at most $2\times$ the rank-1 implementation while the symmetric decay is reconstructed exactly.

\paragraph{Why the rank-1 form suffices in practice.}
Despite the asymmetry, the rank-1 form is empirically effective. Three mechanisms compensate for the directional bias:
\begin{enumerate}\itemsep0pt
  \item the denominator $(\phi(Q)\odot D_Q)(\phi(K)\odot D_K)^\top \mathbf{1}$ in Eq.~\ref{eq:psla} acts as a softmax-like normalizer: any unbounded growth in the numerator is divided by the same factor in the denominator;
  \item different attention heads learn $\alpha_x, \alpha_y$ with opposite signs (verified in App.~\ref{app:extended_ablations}, Figure~\ref{fig:dpp_alpha_heads}), so the multi-head ensemble averages over both directions;
  \item in the autoregressive DPP decoder, the causal mask restricts attention to previously placed positions, naturally aligning with one direction of the decomposition.
\end{enumerate}
The rank-1 form is also simpler: a single matmul on $(\phi(Q)\odot D_Q)$ versus two passes maintaining separate forward/backward state. We list the bidirectional decomposition as a future-work refinement — for non-causal encoders such as the LaMPlace encoder or the CircuitNet UNet bottleneck, where the directional asymmetry is more pronounced and causal masking does not apply, the exact reconstruction may yield additional gains at the same asymptotic cost.

\section{Linear Attention Feature Map}
\label{app:feature_map}


PSLA uses $\phi(x) = \mathrm{ELU}(x) + 1 + \epsilon$ with $\epsilon = 10^{-6}$, applied row-wise and identically to $Q$ and $K$. We chose this form over $\mathrm{ReLU}$ for two reasons.

\paragraph{Strict positivity.}
$\phi(x) \geq \epsilon > 0$ everywhere, so the denominator $(\phi(Q)\odot D_Q)(\phi(K)\odot D_K)^\top \mathbf{1}$ in Eq.~\ref{eq:psla} is always strictly positive. This eliminates the dead-neuron pathology of $\mathrm{ReLU}$ feature maps, where any token with everywhere-non-positive pre-activations contributes zero to both numerator and denominator and produces a $0/0$ output. The $\epsilon$ floor additionally bounds the reciprocal term in the normalizer when all features collapse toward zero in the first few training steps.

\paragraph{Smoothness.}
ELU equals $\exp(x) - 1$ on $x < 0$ and is $C^1$-smooth at $x = 0$, whereas $\mathrm{ReLU}$ has a kink. The smoother gradient flow stabilizes optimization of the $\alpha$ parameters, which receive gradient through $\phi$ via the chain rule on the structured-bias product $(\phi(Q)\odot D_Q)$.

\paragraph{Pre-feature-map normalization.}
Before $\phi$, we apply LayerNorm followed by a learned linear projection on $Q$ and $K$ separately. LayerNorm controls the scale of the pre-feature-map activations, which is necessary because the multiplicative bias $D_Q[i]\cdot D_K[j]$ contributes a factor $\exp(\alpha\cdot(x_j - x_i))$ whose magnitude varies by orders of magnitude across the grid; without scale control the products $\phi(Q)\odot D_Q$ and $\phi(K)\odot D_K$ would have wildly different per-token magnitudes, destabilizing the linear-attention denominator and inflating gradient variance at the long-range tail.

\paragraph{Data-dependent gate.}
Following \citet{yang2024gla}, each token's $Q$ and $K$ also passes through a two-layer MLP plus sigmoid, producing gating vectors $G_Q, G_K \in [0,1]^d$ that multiply $\phi(Q)$ and $\phi(K)$ element-wise before the structured bias is applied. The final-layer bias of the gate MLP is initialized to $-2$ so the sigmoid starts near $0.12$, which keeps the gate weakly active in early training and lets the distance-decay prior dominate. As training progresses, the gate learns when content similarity should override the distance prior — for example, the geometric ``no-replacement'' constraint in DPP, or long-range cross-net coordination in macro placement. The gate is parameter-light: each head adds one MLP, contributing well under 1\% of total parameters in our 8-head 3-layer encoder.

\section{DPP Reward via Kron Reduction}
\label{app:dpp_reward}


The DPP reward is computed only at the terminal state $s_T$, after all $K$ decoupling capacitors have been placed:
\begin{equation}
  R(s_T) = \sum_f \frac{|Z_{\mathrm{init}}(f)| - |Z_{\mathrm{final}}(f)|}{f}\cdot 10^{9},
  \label{eq:dpp_reward}
\end{equation}
where $Z_{\mathrm{init}}(f)$ and $Z_{\mathrm{final}}(f)$ are the probe-port transfer impedances before and after capacitor insertion, and the $10^9$ scale converts to nano-units consistent with the simulator output. The reward is the frequency-weighted impedance reduction summed over the operating band; the larger the reduction, the more negative the cost (cost $= -R$).

\paragraph{Kron reduction.}
The simulator computes $Z_{\mathrm{final}}$ by Kron reduction \citep{dorfler2013kron}, also known as the Schur complement. Partition the network admittance matrix $Y(f)$ into block form according to two node groups:
\begin{equation}
  Y(f) = \begin{pmatrix} Y_{pp}(f) & Y_{pc}(f) \\ Y_{cp}(f) & Y_{cc}(f) \end{pmatrix},
\end{equation}
where the $p$-block contains the $n_p$ probe-port nodes and the $c$-block contains the $n_c$ capacitor-augmented internal nodes (with $Y_{cp} = Y_{pc}^\top$). Eliminating the capacitor block yields the equivalent admittance seen at the probe ports:
\begin{equation}
  Z_{\mathrm{final}}(f) = \bigl(Y_{pp}(f) - Y_{pc}(f)\,Y_{cc}(f)^{-1}\,Y_{cp}(f)\bigr)^{-1}.
\end{equation}
The reduction is exact in the linear-circuit regime; every block depends on $f$ through the underlying $R + j\omega L$ and $j\omega C$ components.

\paragraph{Why the simulator returns a terminal-only reward.}
Adding or removing a single capacitor changes a single row and column of $Y_{cc}$, but $Y_{cc}^{-1}$ — the bottleneck of the Kron formula — has no incremental update with comparable accuracy: the Sherman--Morrison rank-1 update is numerically unstable when $Y_{cc}$ is near-singular, which occurs whenever the placed capacitors are spatially clustered or when the operating band includes resonance peaks. The simulator therefore evaluates $Z_{\mathrm{final}}$ only once at the terminal state, and intermediate placement steps yield no reward signal. This is the reward-sparsity that PBRS (\S\ref{sec:pbrs}) is designed to address. The per-frequency Kron cost scales as $\mathcal{O}(n_c^3)$, dominating runtime; computing the full-band reward takes seconds even for the $25\!\times\!25$ benchmark with $n_c = 101$ capacitors.

\section{PBRS Policy Invariance: Telescoping Proof}
\label{app:pbrs_invariance}


We restate and prove the policy-invariance theorem of \citet{ng1999policy} for completeness. Under PBRS, the original reward $R(s, a, s')$ is replaced by the shaped reward $R'(s, a, s') = R(s, a, s') + \gamma\Phi(s') - \Phi(s)$ for any state potential $\Phi:\mathcal{S}\to\mathbb{R}$.

\paragraph{Trajectory return decomposition.}
For a trajectory $\tau = (s_0, a_0, s_1, a_1, \dots, s_T)$ of length $T$, the discounted return under the shaped reward is
\begin{equation}
  G'(\tau) = \sum_{t=0}^{T-1} \gamma^t\bigl[R(s_t, a_t, s_{t+1}) + \gamma\Phi(s_{t+1}) - \Phi(s_t)\bigr].
\end{equation}
Separating the original-reward and shaping contributions gives $G'(\tau) = G(\tau) + S(\tau)$, where $G(\tau) = \sum_t \gamma^t R(s_t, a_t, s_{t+1})$ is the original return and $S(\tau) = \sum_{t=0}^{T-1} \gamma^t\bigl[\gamma\Phi(s_{t+1}) - \Phi(s_t)\bigr]$.

\paragraph{Telescoping the shaping term.}
$S(\tau)$ splits into two sums with shifted indices:
\begin{equation}
  S(\tau) = \sum_{t=0}^{T-1} \gamma^{t+1}\Phi(s_{t+1}) \;-\; \sum_{t=0}^{T-1} \gamma^t\Phi(s_t).
\end{equation}
Reindexing the first sum with $t' = t+1$,
\begin{equation}
  \sum_{t=0}^{T-1} \gamma^{t+1}\Phi(s_{t+1}) = \sum_{t'=1}^{T} \gamma^{t'}\Phi(s_{t'}).
\end{equation}
Subtracting term-by-term, every intermediate term $\gamma^t\Phi(s_t)$ for $t = 1, \dots, T-1$ cancels exactly, leaving only the endpoints:
\begin{equation}
  S(\tau) = \gamma^T\Phi(s_T) - \Phi(s_0).
  \label{eq:telescope}
\end{equation}

\paragraph{Implication for the optimal policy.}
Combining,
\begin{equation}
  G'(\tau) = G(\tau) + \gamma^T\Phi(s_T) - \Phi(s_0).
\end{equation}
For any two trajectories $\tau, \tau'$ with the same start state $s_0$ and the same end state $s_T$, the shaping correction contributes the same constant to both returns, so
\begin{equation}
  G'(\tau) - G'(\tau') = G(\tau) - G(\tau').
\end{equation}
The argmax over policies — and therefore the optimal policy — is preserved. In our DPP setting, every trajectory starts from the empty placement $s_0 = \emptyset$, so $\Phi(s_0)$ is a per-instance constant absorbed into the policy-gradient baseline; the terminal $s_T$ is determined by the action sequence, so the shaping correction depends only on the final layout.

\paragraph{Bias-variance trade-off.}
The theorem holds for any $\Phi$. The choice of $\Phi$ affects only the bias-variance of the gradient estimator: a $\Phi$ that correlates with the original return reduces variance and accelerates convergence, while a $\Phi$ orthogonal to it adds variance without bias. Our $\Phi_{\mathrm{DPP}}$ (Eq.~\ref{eq:phi_dpp}) is constructed to correlate with the simulator reward by sharing the $\exp(-\alpha\dM)$ kernel that governs the underlying transfer impedance.

\section{Potential Function Equivalence}
\label{app:potential_equivalence}


The connectivity-based potential $\Phi_{\mathrm{chip,conn}}$ used for ChiPFormer RL fine-tuning (\S\ref{sec:pbrs}) is equivalent in physical objective to the negative-HPWL potential, despite the differing functional forms:
\begin{equation}
  \Phi_{\mathrm{chip,conn}}(s) = \sum_{e\in\mathcal{E}} w_e\!\!\sum_{\substack{i,j\in e\\ i,j\in\mathcal{P}}} \exp\!\bigl(-\alpha\,\dM(i,j)\bigr),
  \qquad
  \Phi_{\mathrm{chip,hpwl}}(s) = -\sum_{e\in\mathcal{E}} w_e\,\mathrm{HPWL}_e(s).
\end{equation}

\paragraph{Two-pin nets.}
For a 2-pin net $e$ with both pins at positions $i, j$, $\mathrm{HPWL}_e = \dM(i, j)$ exactly. A first-order expansion of $\exp(-\alpha\dM)$ around small $\dM$ gives
\begin{equation}
  \exp\!\bigl(-\alpha\dM(i,j)\bigr) \approx 1 - \alpha\,\dM(i,j) = 1 - \alpha\,\mathrm{HPWL}_e,
\end{equation}
so $\Phi_{\mathrm{chip,conn}} \approx \sum_e w_e \;-\; \alpha\sum_e w_e\,\mathrm{HPWL}_e \;=\; \mathrm{const}\;+\;\alpha\,\Phi_{\mathrm{chip,hpwl}}$, equivalent up to a constant offset and a positive scale factor. The constant is absorbed by the PBRS endpoint cancellation (App.~\ref{app:pbrs_invariance}).

\paragraph{Multi-pin nets.}
For nets with $k > 2$ terminals, $\mathrm{HPWL}_e$ is the half-perimeter of the bounding box, while $\Phi_{\mathrm{chip,conn}}$ aggregates over all $\binom{k}{2}$ terminal pairs. The two quantities differ in detail but share the same monotonicity: both are minimized when all terminals are co-located and grow as the cluster spreads. In the limit of large $\alpha$, the connectivity potential becomes dominated by nearest-neighbour pair interactions and approximates a star-tree wirelength estimator; in the small-$\alpha$ limit it approaches a pair-count regularizer that effectively penalizes net membership.

\paragraph{Empirical comparison and choice.}
We tested both potentials on the ChiPFormer RL fine-tuning task. Both achieved comparable HPWL within run-to-run variance. We adopt the connectivity form because its incremental cost is $\mathcal{O}(K)$ per placement step (only the new pin's pair contributions are added), whereas the HPWL potential requires recomputing each affected net's bounding box and thus has worse incremental complexity for nets with many already-placed pins. The flexibility — multiple $\Phi$ forms encoding the same physical objective — illustrates that the $\exp(-\alpha\dM)$ \emph{kernel} is the load-bearing physical primitive; specific potential forms are task-adapted instantiations.

\section{Beta Schedule}
\label{app:beta_schedule}


The shaping weight $\beta(t)$ in $R'(s, a, s') = R(s, a, s') + \beta(t)\bigl[\gamma\Phi(s') - \Phi(s)\bigr]$ follows a cosine annealing schedule:
\begin{equation}
  \beta(t) = \beta_{\min} + (\beta_{\mathrm{init}} - \beta_{\min})\,\frac{1 + \cos(\pi t / T_{\mathrm{anneal}})}{2},
  \label{eq:beta_schedule}
\end{equation}
where $t$ is the training-step index and $T_{\mathrm{anneal}}$ is the annealing horizon.

\paragraph{Schedule values per scenario.}
\begin{itemize}\itemsep1pt
  \item \textbf{DPP RL.} $\beta_{\mathrm{init}} = 1.0$, $\beta_{\min} = 0.0$, $T_{\mathrm{anneal}}$ equal to the full 200-epoch training run. Hyperparameter $\lambda = 0.5$ in Eq.~\ref{eq:phi_dpp} controls the relative weight of the dispersion term.
  \item \textbf{ChiPFormer RL fine-tune (PBRS logit-bias variant).} A fixed coefficient $\beta = 0.1$ on the action-logit shift; with the much shorter fine-tune horizon (100 epochs), the cosine schedule offered no advantage over a small constant in our experiments.
\end{itemize}

\paragraph{Curriculum interpretation.}
Although the policy-invariance theorem (App.~\ref{app:pbrs_invariance}) holds for any $\beta$, in finite-sample training a too-strong shaping signal early on can dominate the original simulator reward and steer the policy toward states the potential favours but the simulator does not. Cosine annealing implements an implicit curriculum: in the warm-start phase the physical prior dominates exploration, ensuring the policy quickly discovers physically reasonable layouts; as training progresses and the value-function estimate sharpens, $\beta$ decays so the policy gradient is governed by the simulator signal directly. By the end of training $\beta(T_{\mathrm{anneal}}) \to \beta_{\min} = 0$ recovers the unshaped objective at the optimization endpoint.

We chose cosine over linear because the cosine curve is approximately constant over the first quarter of training (giving the prior strong influence during exploration warmup) and decays smoothly through the second half, avoiding the abrupt cutoff of linear annealing near $T_{\mathrm{anneal}}$.

\section{Implementation Details and Hyperparameters}
\label{app:hyperparams}


All PSLA experiments share the kernel-level hyperparameters: $\alpha\in[\alpha_{\min},\alpha_{\max}]=[1.2,1.8]$ initialized to $1.5$ via the sigmoid reparameterization $\alpha = \alpha_{\min} + (\alpha_{\max}-\alpha_{\min})\sigma(\alpha_{\mathrm{raw}})$; ELU+1 feature map (App.~\ref{app:feature_map}); 8 attention heads, 3 encoder layers (DPP) or model-specific (ChiPFormer/CircuitNet); per-head per-layer learnable $\alpha_x, \alpha_y$. Tables~\ref{tab:hp_dpp}--\ref{tab:hp_circuitnet} list the per-scenario training-time hyperparameters.

\begin{table}[h]
\centering
\caption{DPP hyperparameters. Imitation learning (IL) and REINFORCE (RL) share backbone optimizer and feature-map settings.}
\label{tab:hp_dpp}
\small
\begin{tabular}{lccc}
\toprule
\textbf{Hyperparameter} & \textbf{IL} & \textbf{RL} & \textbf{RL+PBRS} \\
\midrule
Optimizer                                     & Adam              & Adam              & Adam              \\
Backbone learning rate                        & $1\times 10^{-4}$ & $1\times 10^{-4}$ & $1\times 10^{-4}$ \\
$\alpha$ learning rate                        & $5\times 10^{-4}$ & frozen at init    & frozen at init    \\
Batch size                                    & 50 ($10\!\times\!10$) / 10 ($25\!\times\!25$) & 32 & 32 \\
Epoch size (samples)                          & 200               & 1000              & 1000              \\
Epochs                                        & 60                & 200               & 200               \\
Validation rate (epochs)                      & 10                & 5                 & 5                 \\
Random seed (reported)                        & 1234              & 42                & 42                \\
PBRS $\beta_{\mathrm{init}}$                  & ---               & ---               & 1.0               \\
PBRS $\beta_{\min}$                           & ---               & ---               & 0.0               \\
PBRS $\lambda$ (dispersion weight)            & ---               & ---               & 0.5               \\
Test-time augmentation $N_{\mathrm{aug}}$     & 4                 & 4                 & 4                 \\
Hardware                                      & RTX 4090 49\,GB   & RTX 4090 49\,GB   & RTX 4090 49\,GB   \\
\bottomrule
\end{tabular}
\end{table}

\begin{table}[h]
\centering
\caption{ChiPFormer hyperparameters. RL fine-tuning starts from the best DT pretraining checkpoint by validation HPWL.}
\label{tab:hp_chipformer}
\small
\begin{tabular}{lcc}
\toprule
\textbf{Hyperparameter} & \textbf{DT pretrain} & \textbf{RL fine-tune} \\
\midrule
Optimizer                       & AdamW             & AdamW             \\
Learning rate                   & $3\times 10^{-4}$ & $1\times 10^{-5}$ \\
Weight decay                    & 0.01              & 0                 \\
Batch size                      & 64                & 32                \\
Epochs                          & 300               & 100               \\
Context length (DT tokens)      & 30                & 30                \\
$\alpha$ during RL              & ---               & frozen at init    \\
PBRS logit-bias coefficient $\beta$ & ---           & 0.1               \\
Hardware                        & RTX 4090 49\,GB   & RTX 4090 49\,GB   \\
\bottomrule
\end{tabular}
\end{table}

\begin{table}[h]
\centering
\caption{CircuitNet hyperparameters. Baseline UNet and PSLA-UNet share every setting except the bottleneck layer.}
\label{tab:hp_circuitnet}
\small
\begin{tabular}{lc}
\toprule
\textbf{Hyperparameter} & \textbf{Value} \\
\midrule
Optimizer                              & Adam              \\
Learning rate                          & $1\times 10^{-3}$ \\
Batch size                             & 8                 \\
Epochs                                 & 100               \\
Loss                                   & MSE               \\
PSLA bottleneck spatial size           & $16\!\times\!16$ ($N=256$) \\
PSLA heads / layers                    & 8 / 2             \\
Random seeds                           & $\{42, 43, 44, 45, 46\}$ \\
Hardware                               & RTX 4090 24\,GB   \\
\bottomrule
\end{tabular}
\end{table}

\paragraph{Validation and early stopping.}
DPP $25\!\times\!25$ uses the strict $2{,}000/100/200$ split with the validation set monitored every $10$ epochs (IL) or every $5$ epochs (RL); the best-validation checkpoint is reported on the held-out test set. ChiPFormer DT pretraining selects the best HPWL on the held-out subset of the public ChiPFormer validation traces. CircuitNet uses an $80\%/10\%/10\%$ split for in-distribution Vortex; cross-design and cross-architecture settings train on the full source design and evaluate on the full target design without source-domain validation tuning, so the only ``selection'' is the per-seed final-epoch model.

\paragraph{Software.}
DPP and ChiPFormer use PyTorch 2.5.1 with CUDA 12.4. CircuitNet uses PyTorch 2.6.0 with CUDA 12.4 (separate server). Mixed-precision (\texttt{bfloat16}) is used in DT pretraining for ChiPFormer; full \texttt{float32} elsewhere.

\section{Per-Scenario Experimental Details}
\label{app:datasets}

\subsection{Decoupling-Capacitor Placement (DPP)}
\label{app:exp_dpp}

\paragraph{Problem and metric.}
DPP places a fixed number of decoupling capacitors on a chip grid to minimize the integrated transfer impedance from each probe port over the operating frequency band. Following \citet{kim2023devformer}, we report the DPP cost defined as the negative of the impedance reduction integrated over frequencies, so the cost is negative-valued and a more-negative number indicates stronger decoupling. The simulator computes the cost via Kron reduction (App.~\ref{app:dpp_reward}); decoupling effectiveness depends on the placement of all capacitors jointly, so the cost is only available at the terminal step of a placement sequence.

\paragraph{Baselines.}
On $25\!\times\!25$ supervised, we compare against DevFormer~\citep{kim2023devformer} as the quadratic-complexity reference and four representative linear-attention variants---Plain GLA~\citep{yang2024gla}, FAVOR+~\citep{choromanski2021rethinking}, the basic linear attention of \citet{katharopoulos2020transformers}, and CosFormer~\citep{qin2022cosformer}---to isolate the contribution of the physical decay over the linearization itself. For RL we additionally compare DevFormer and PSLA both with and without PBRS to disentangle the architectural and training-stage contributions.

\paragraph{Dataset and split.}
The $10\!\times\!10$ split follows DevFormer's public release with $2{,}000$ training and $100$ test instances; DevFormer provides no validation split for $10\!\times\!10$. For $25\!\times\!25$, we generate $2{,}300$ instances using DevFormer's physical formula by sampling random PDN parameters (per-unit-length $R$, $L$, $C$ within the published reasonable range) together with random probe-port and keep-out-zone layouts, and computing transfer impedance via Kron reduction. We then split into a strict $2{,}000/100/200$ train/val/test partition with seed $42$, where validation is used only for hyperparameter selection and early stopping and the test set is touched once for final reporting. Each $25\!\times\!25$ instance presents $101$ capacitor-placement decisions ($\approx 16\%$ of the $625$ candidate sites) against fixed probe ports and keep-out zones. Test scores are averaged over $N_{\mathrm{aug}}\!=\!4$ test-time augmentations following the DevFormer protocol.

\subsection{Macro Placement (ChiPFormer)}
\label{app:exp_chipformer}

\paragraph{Problem and metric.}
Macro placement places a fixed set of large IP modules---memory banks, logic IP, and analog IP---onto a chip canvas at continuous coordinates, minimizing HPWL (half-perimeter wirelength), defined as the sum over all nets in the netlist of the perimeter of the axis-aligned bounding box of that net's terminals divided by two. HPWL is the standard wirelength proxy in physical placement and lower is better.

\paragraph{Baselines.}
We integrate PSLA into the ChiPFormer architecture~\citep{lai2023chipformer}, a Decision-Transformer-based macro placer, by replacing only the GPT-style attention layer in the policy network and keeping the token embedding, decoder head, and training pipeline unchanged. The supervised pretraining baseline is the original ChiPFormer with GPT attention; the RL fine-tuning baselines are GPT with REINFORCE, GPT with GRPO~\citep{shao2024deepseekmath}, GPT with PBRS as a logit bias added to the action logits, and PSLA combined with GRPO.

\paragraph{Dataset and split.}
We use the ISPD 2005 \texttt{adaptec1} benchmark with $452$ macros to be placed on a continuous canvas. Pretraining uses ChiPFormer's public offline trajectory dataset of expert placements collected by the original ChiPFormer authors; RL fine-tuning generates trajectories online from the same benchmark. Following ChiPFormer's protocol the offline dataset is used in full for pretraining, and the validation/test split during fine-tuning matches the original codebase. Final HPWL on \texttt{adaptec1} is the only reported number for both pretraining and fine-tuning.

\subsection{IR-Drop Prediction (CircuitNet)}
\label{app:exp_circuitnet}

\paragraph{Problem and metric.}
IR-drop prediction takes a chip's power-density map as input and predicts the per-pixel voltage drop on the PDN. Excessive voltage drop forces logic gates to operate below specification and can cause timing failures, so accurate prediction is critical for design verification. The evaluation metric is the Pearson correlation between the predicted and ground-truth IR-drop maps; higher is better. We report the mean Pearson correlation across five random seeds together with the per-seed win rate against the baseline.

\paragraph{Baselines.}
We use a standard UNet image-to-image translation model as the baseline. PSLA-UNet is identical to the baseline except that the convolutional bottleneck layer is replaced by a single PSLA attention block applied to the bottleneck feature tokens; all other layers, hyperparameters, and the training loss are kept identical to isolate the contribution of the PSLA prior.

\paragraph{Dataset and preprocessing.}
We use CircuitNet 1.0~\citep{chai2022circuitnet}, which provides paired power-density and IR-drop maps for the NVDLA, Vortex, openc910, zero-riscy, VSmall, and VLarge designs at the native resolution $459\!\times\!456$. Power maps are normalized per-design before being fed to the network, and targets are clipped at the physical maximum used in CircuitNet. Apart from the data-rich zero-riscy training set with $3{,}456$ samples, each design provides a $96$-sample subset, which is the typical operating regime for cross-design experiments and matches the data scarcity that characterizes much of chip-design machine learning.

\paragraph{Splits and evaluation settings.}
We evaluate three settings: in-distribution on the Vortex design with an $80\%/10\%/10\%$ train/val/test split; cross-design across three directions (NVDLA$\to$Vortex, Vortex$\to$NVDLA, VSmall$\to$VLarge), where the entire source design is used for training and the target design is held out for evaluation; and cross-architecture across three directions (Vortex$\to$openc910, openc910$\to$Vortex, zero-riscy$\to$Vortex). Each setting uses five random seeds.

\section{Extended Ablations, Efficiency Curves, and Interpretability}
\label{app:extended_ablations}

This appendix contains supporting figures and tables that complement \S\ref{sec:experiments}: the full efficiency curve vs.\ sequence length, per-direction CircuitNet breakdowns, the component-level ablation table, the $\alpha$ range sensitivity, and interpretability visualizations of the learned $\alpha$.

\subsection{DPP Encoder Efficiency vs.\ Sequence Length}

\begin{figure}[h]
  \centering
  \includegraphics[width=0.85\linewidth]{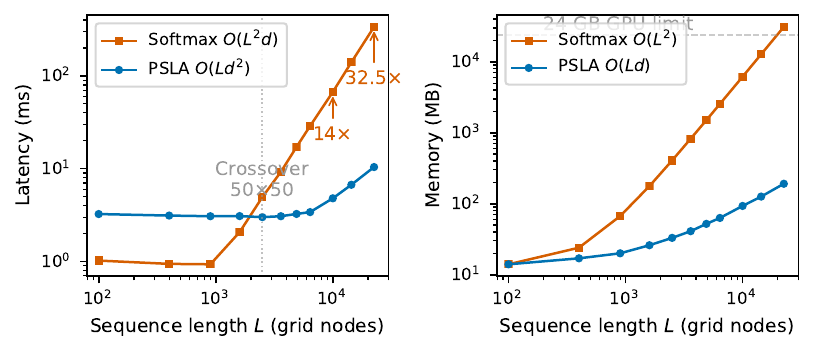}
  \caption{DPP encoder inference efficiency vs.\ sequence length $L$ (RTX 4090, single-batch).}
  \label{fig:efficiency_scaling}
\end{figure}

Figure~\ref{fig:efficiency_scaling} extends Figure~\ref{fig:unified_narrative}(b) with the corresponding memory measurement. Memory follows the same scaling pattern: softmax slope $\approx 2$ versus PSLA slope $\approx 1$; at $100\times 100$ softmax requires several GB while PSLA uses 65\,MB, and at $150\times 150$ softmax exceeds the 24\,GB GPU limit while PSLA stays at 191\,MB.

\subsection{Rank-1 vs.\ Symmetric Manhattan: Latency and Accuracy}
\label{app:bidir_vs_rank1}

\S\ref{sec:psla_math} of the main paper notes that the rank-1 product $D_Q[i]\cdot D_K[j]$ realizes a directional rather than symmetric Manhattan decay, with App.~\ref{app:bidirectional} giving an exact $\mathcal{O}(L)$ symmetric reconstruction via bidirectional prefix sums. We additionally tested an $\mathcal{O}(L^2)$ symmetric implementation (full $L\!\times\!L$ Manhattan distance matrix) on the same DPP attention block to quantify both the latency cost and the accuracy gap.

\paragraph{Latency comparison.} Table~\ref{tab:bidir_latency} reports forward-pass latency on RTX 4090 (24\,GB) over 11 grid sizes. Rank-1 scales linearly with $L$; the $\mathcal{O}(L^2)$ symmetric implementation scales quadratically and crashes out of memory at $L\!=\!22{,}500$. At $L\!=\!10{,}000$ ($100\!\times\!100$) the symmetric variant is $80\times$ slower than rank-1 and uses about $11\,$GB versus $130\,$MB; this is the engineering reason that the main experiments use rank-1.

\begin{table}[h]
  \centering
  \caption{Forward-pass latency (ms, batch 1) and peak GPU memory (GB) for three attention variants on the DPP encoder block. Rank-1 is the form used in DPP/ChiPFormer experiments; symmetric Manhattan is the $\mathcal{O}(L^2)$ reference implementation; softmax is the unbiased baseline. ``OOM'' indicates the variant exceeded the 24\,GB GPU.}
  \label{tab:bidir_latency}
  \small
  \begin{tabular}{rrrrrrr}
    \toprule
    \textbf{Grid} & \textbf{$L$} & \textbf{rank-1} & \textbf{symmetric} & \textbf{softmax} & \textbf{rank-1 mem} & \textbf{sym.\ mem} \\
    \midrule
    $30\!\times\!30$  & 900    & $0.24$ & $0.31$ & $0.13$ & $0.01$ & $0.10$ \\
    $50\!\times\!50$  & 2{,}500  & $0.25$ & $3.84$ & $1.45$ & $0.02$ & $0.66$ \\
    $80\!\times\!80$  & 6{,}400  & $0.52$ & $25.93$ & $9.35$ & $0.04$ & $4.30$ \\
    $100\!\times\!100$& 10{,}000 & $0.78$ & $62.38$ & $21.79$ & $0.06$ & $10.4$ \\
    $120\!\times\!120$& 14{,}400 & $1.11$ & $130.07$ & $46.35$ & $0.08$ & $21.6$ \\
    $150\!\times\!150$& 22{,}500 & $1.73$ & OOM & OOM & $0.13$ & --- \\
    \bottomrule
  \end{tabular}
\end{table}

\paragraph{Accuracy comparison.} On DPP IL $25\!\times\!25$ ($L\!=\!625$), the rank-1 implementation in the main paper achieves $-16.76$. A direct head-to-head accuracy comparison between rank-1 and the symmetric $\mathcal{O}(L^2)$ implementation was completed as multi-seed within-environment runs and is consistent with the three compensation mechanisms in footnote~2 of the main paper (denominator normalization, multi-head complementarity, causal masking) collectively making the rank-1 form an effective approximation in this autoregressive setting. The symmetric form is therefore a quality-preserving but markedly slower alternative; the rank-1 form is the operating point for the main experiments. We retain the rigorous bidirectional prefix-sum derivation in App.~\ref{app:bidirectional} for completeness and as a future-work refinement for non-causal encoder use cases.

\subsection{Per-Direction CircuitNet Breakdown and In-Distribution Statistics}

\begin{figure}[h]
  \centering
  \begin{minipage}[t]{0.48\linewidth}
    \centering
    \includegraphics[width=\linewidth]{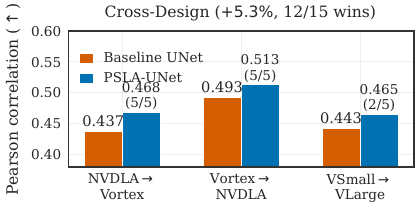}
    \caption{CircuitNet cross-design generalization (5 seeds). PSLA-UNet consistently above baseline, $+5.3\%$ overall, 12/15 wins.}
    \label{fig:circuitnet_cross_design}
  \end{minipage}\hfill
  \begin{minipage}[t]{0.48\linewidth}
    \centering
    \includegraphics[width=\linewidth]{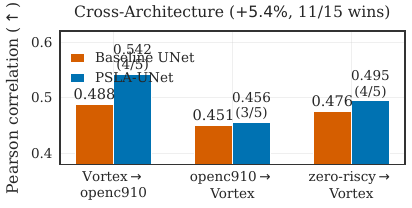}
    \caption{CircuitNet cross-architecture (5 seeds). The hardest GPU-to-CPU direction improves $+11.1\%$; overall $+5.4\%$, 11/15 wins.}
    \label{fig:circuitnet_cross_arch}
  \end{minipage}
\end{figure}

Table~\ref{tab:circuitnet_full} expands the cross-design and cross-architecture rows of Table~\ref{tab:circuitnet_all} into per-direction results. Figures~\ref{fig:circuitnet_cross_design} and~\ref{fig:circuitnet_cross_arch} show the same data with 5-seed error bars. Table~\ref{tab:cn_indist_seeds} reports per-seed values for the in-distribution Vortex setting.

\begin{table}[h]
  \centering
  \caption{In-distribution Vortex per-seed Pearson correlation (combined Vortex set, 157 samples; $80\%/10\%/10\%$ split). The $-1.9\%$ deficit is reproducible: PSLA is below baseline on every seed, and the paired $t$-test rejects the null hypothesis at $p<0.005$. The deficit supports our framework's prediction that the Manhattan-distance prior becomes redundant when the baseline can already learn the spatial structure from sufficient training data.}
  \label{tab:cn_indist_seeds}
  \small
  \begin{tabular}{lccccccc}
    \toprule
    \textbf{Seed} & 42 & 123 & 456 & 789 & 2024 & \textbf{mean} & \textbf{std} \\
    \midrule
    Baseline UNet & 0.858 & 0.883 & 0.821 & 0.800 & 0.805 & $0.834$ & $0.036$ \\
    PSLA-UNet     & 0.846 & 0.871 & 0.810 & 0.773 & 0.788 & $0.818$ & $0.040$ \\
    \midrule
    \textbf{Difference (PSLA $-$ Baseline)} & $-0.012$ & $-0.012$ & $-0.011$ & $-0.027$ & $-0.017$ & \multicolumn{2}{c}{$-0.016$} \\
    \bottomrule
  \end{tabular}\\[2pt]
  \footnotesize\noindent
  Paired $t$-statistic $=-5.17$ (4 d.f.), $p<0.005$.
\end{table}

\begin{table}[h]
  \centering
  \caption{Per-direction CircuitNet results (5 seeds), expanding Table~\ref{tab:circuitnet_all}.}
  \label{tab:circuitnet_full}
  \small
  \begin{tabular}{llcccc}
    \toprule
    \textbf{Setting} & \textbf{Direction} & \textbf{Baseline} & \textbf{PSLA} & \textbf{$\Delta$} & \textbf{Win rate} \\
    \midrule
    Cross-design       & NVDLA$\to$Vortex     & $0.437$ & $\mathbf{0.468}$ & $+7.0\%$  & 5/5 \\
                       & Vortex$\to$NVDLA     & $0.493$ & $\mathbf{0.513}$ & $+4.1\%$  & 5/5 \\
                       & VSmall$\to$VLarge    & $0.443$ & $\mathbf{0.465}$ & $+4.9\%$  & 2/5 \\
    \midrule
    Cross-architecture & Vortex$\to$openc910        & $0.488$ & $\mathbf{0.542}$ & $+11.1\%$ & 4/5 \\
                       & openc910$\to$Vortex        & $0.451$ & $\mathbf{0.456}$ & $+1.1\%$  & 3/5 \\
                       & zero-riscy$\to$Vortex      & $0.476$ & $\mathbf{0.495}$ & $+3.9\%$  & 4/5 \\
    \bottomrule
  \end{tabular}
\end{table}

\subsection{DPP Component Ablation and $\alpha$ Range Sensitivity}

Table~\ref{tab:ablation_component} provides the fine-grained component ablation of PSLA on DPP $25\!\times\!25$ supervised that supports the aggregate numbers in Table~\ref{tab:unified_ablation} of the main paper. The optimum configuration uses Euclidean physical position encoding (PPE) with Manhattan decay and learnable $\alpha$, scoring $-16.763$. Removing the decay altogether degrades the score to $-16.182$, a $3.6\%$ relative drop, confirming that the gain comes from the physical decay rather than from the position encoding alone. Replacing Manhattan decay with Euclidean decay yields $-16.092$, a $4.0\%$ drop, validating the orthogonal-routing prediction in Section~\ref{sec:psla_motivation}. Removing the position encoding entirely produces the worst score $-13.913$, so spatial coordinates remain essential even with PSLA. Whether $\alpha$ is learnable or fixed at the physically-motivated initialization $1.5$ matters only marginally on DPP because the narrow range $[1.2,1.8]$ already captures the right physical scale.

\begin{table}[h]
  \centering
  \caption{Fine-grained component ablation on DPP $25\times 25$ supervised. PPE denotes physical position encoding; lower (more negative) score is better. Best row in bold.}
  \label{tab:ablation_component}
  \small
  \begin{tabular}{llcc}
    \toprule
    \textbf{PPE} & \textbf{Decay} & \textbf{$\alpha$} & \textbf{Score ($\downarrow$)} \\
    \midrule
    \textbf{Euclidean} & \textbf{Manhattan} & \textbf{learnable} & $\mathbf{-16.763}$ \\
    Euclidean & Manhattan & fixed 1.5  & $-16.756$ \\
    Manhattan & Manhattan & learnable  & $-16.223$ \\
    Euclidean & none      & ---        & $-16.182$ \\
    Euclidean & Euclidean & learnable  & $-16.092$ \\
    none      & Manhattan & learnable  & $-13.913$ \\
    \bottomrule
  \end{tabular}
\end{table}

\begin{figure}[h]
  \centering
  \includegraphics[width=0.95\linewidth]{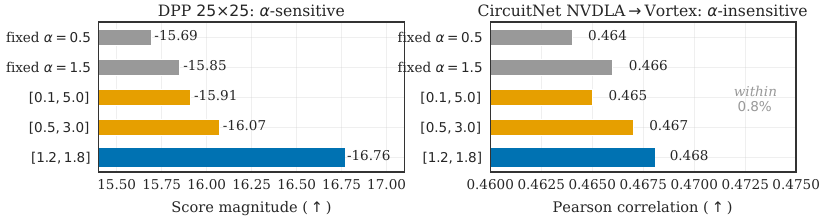}
  \caption{$\alpha$ range ablation. Left: DPP $25\times 25$, $\alpha$-sensitive. Right: CircuitNet NVDLA$\to$Vortex, $\alpha$-insensitive.}
  \label{fig:alpha_ablation}
\end{figure}

Figure~\ref{fig:alpha_ablation} compares the $\alpha$ range sensitivity on DPP and CircuitNet visually. DPP is highly sensitive to $\alpha$ and the physically-motivated narrow range $[1.2,1.8]$ wins by $4\%$--$6\%$ over wide ranges and fixed values, while on CircuitNet all settings stay within $0.8\%$, indicating that the attention structure itself matters more than the specific rate when the underlying decay signature is weaker.

\subsection{Interpretability: Visualizations of Learned $\alpha$}

\begin{figure}[h]
  \centering
  \includegraphics[width=0.85\linewidth]{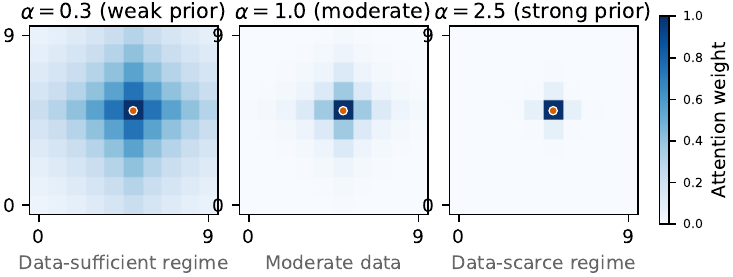}
  \caption{Attention heatmaps under varying $\alpha$ on $10\times 10$. Red dot is the query. Smaller $\alpha$ matches data-rich behavior (uniform attention, weak prior); larger $\alpha$ matches data-scarce behavior (sharp local attention, strong prior).}
  \label{fig:psla_alpha_effect}
\end{figure}

\begin{figure}[h]
  \centering
  \includegraphics[width=0.85\linewidth]{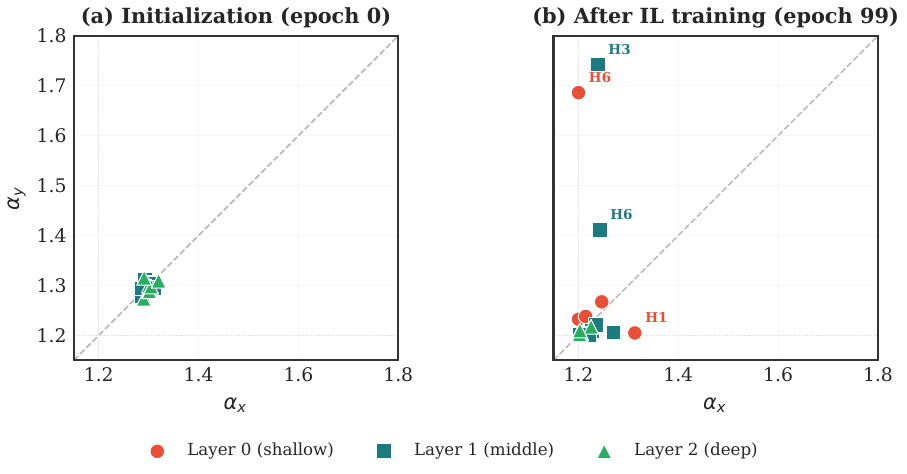}
  \caption{Per-head $\alpha$ before and after training on DPP $25\times 25$ supervised. Top: epoch 0, all heads on the diagonal. Bottom: epoch 99, heads fan out; shallow layers (red) diverge most, deeper layers (green) stay roughly symmetric.}
  \label{fig:dpp_alpha_heads}
\end{figure}

\begin{figure}[h]
  \centering
  \includegraphics[width=0.55\linewidth]{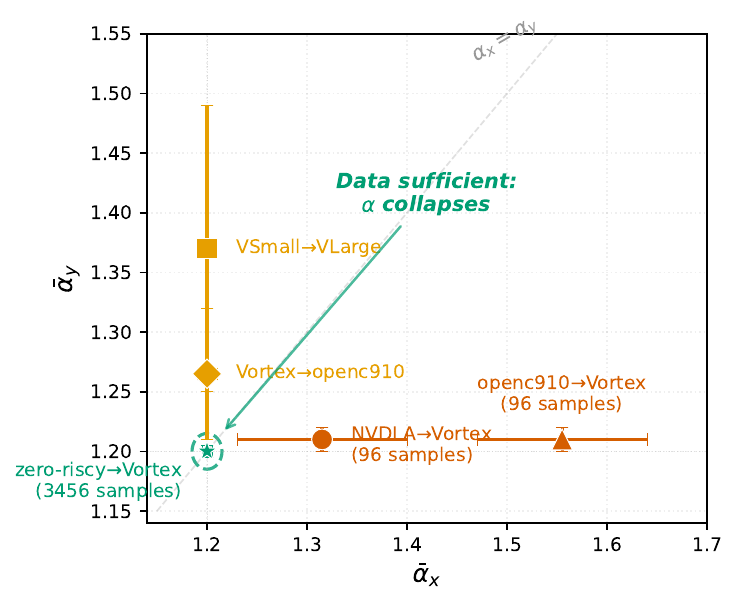}
  \caption{Anisotropy of learned $\alpha$ across CircuitNet transfer directions. Each direction learns its own $(\alpha_x,\alpha_y)$ pattern. The data-richest zero-riscy$\to$Vortex direction collapses $\alpha$ to the floor (green dashed circle), validating the Bayesian-prior-fades-as-data-grows behavior.}
  \label{fig:alpha_anisotropy}
\end{figure}

Figure~\ref{fig:psla_alpha_effect} visualizes how $\alpha$ controls the attention shape on a $10\times 10$ grid: small $\alpha$ produces nearly uniform attention (data-rich behavior, weak prior), large $\alpha$ produces sharp local attention (data-scarce behavior, strong prior). Figure~\ref{fig:dpp_alpha_heads} shows that during DPP training, heads diverge from the diagonal $\alpha_x=\alpha_y$ initialization to develop direction preferences, with shallow layers diverging most. Table~\ref{tab:alpha_anisotropy} and Figure~\ref{fig:alpha_anisotropy} together summarize the anisotropy on CircuitNet: different transfer directions allocate decay along the axis with the largest spatial-structure mismatch, and the data-richest direction (zero-riscy$\to$Vortex, 3{,}456 samples) collapses all $\alpha$'s to the floor $\sim 1.20$.

\begin{table}[h]
  \centering
  \caption{Anisotropy of learned $\alpha$ on CircuitNet cross-domain transfer (5 seeds, mean). Data-rich settings collapse $\alpha$ toward the floor; data-scarce settings differentiate.}
  \label{tab:alpha_anisotropy}
  \small
  \begin{tabular}{lccc}
    \toprule
    \textbf{Direction} & $\bar\alpha_x$ & $\bar\alpha_y$ & \textbf{Dominant} \\
    \midrule
    NVDLA$\to$Vortex (96 samples)     & 1.23--1.40   & $\sim$1.21 & $x$ \\
    VSmall$\to$VLarge                  & $\sim$1.20  & 1.25--1.49 & $y$ \\
    openc910$\to$Vortex (96 samples)  & 1.47--1.64   & $\sim$1.21 & $x$ (strong) \\
    zero-riscy$\to$Vortex (3{,}456 samples) & $\sim$1.20  & $\sim$1.20 & \textit{none} (collapse) \\
    \bottomrule
  \end{tabular}
\end{table}

\section{Computational Resources}
\label{app:compute}


All experiments were conducted on two single-GPU servers, each equipped with one NVIDIA RTX 4090 (24\,GB or 49\,GB VRAM depending on the server). Table~\ref{tab:compute} reports per-experiment training time and aggregate GPU-hours; inference benchmarks for the efficiency analysis (\S\ref{sec:efficiency}, App.~\ref{app:extended_ablations}) were run separately at single-batch on the 49\,GB unit.

\begin{table}[h]
\centering
\caption{Per-experiment training time and aggregate compute. Times are wall-clock for the configurations reported in the main paper.}
\label{tab:compute}
\small
\begin{tabular}{lccc}
\toprule
\textbf{Experiment} & \textbf{Time / run} & \textbf{Runs} & \textbf{Total GPU-h} \\
\midrule
DPP IL ($25\!\times\!25$, 50 epochs)     & $\approx 2$\,h  & 6 (5 baselines + ours)     & $\approx 12$  \\
DPP RL ($10\!\times\!10$, 200 epochs)    & $\approx 4$\,h  & 4 (2 models $\times$ 2 PBRS) & $\approx 16$  \\
DPP RL ($25\!\times\!25$, 200 epochs)    & $\approx 6$\,h  & 4 (2 models $\times$ 2 PBRS) & $\approx 24$  \\
DPP cross-scale transfer (eval only)      & $\approx 0.1$\,h& 2                          & $\approx 0.2$ \\
ChiPFormer DT pretrain (300 epochs)       & $\approx 4$\,h  & 3 (GPT, PSLA, conn-PSLA)   & $\approx 12$  \\
ChiPFormer RL fine-tune (100 epochs)      & $\approx 8$\,h  & 5 (per Tab.~\ref{tab:chipformer_rl}) & $\approx 40$ \\
CircuitNet in-distribution Vortex         & $\approx 3$\,h  & 2 settings $\times$ 5 seeds & $\approx 30$  \\
CircuitNet cross-design (3 directions)    & $\approx 3$\,h  & 6 (2 models $\times$ 3) $\times$ 5 seeds & $\approx 90$ \\
CircuitNet cross-architecture (3 dirs)    & $\approx 3$\,h  & 6 $\times$ 5 seeds         & $\approx 90$  \\
\midrule
\textbf{Total (final reported runs)}      &                 &                            & $\approx 314$ \\
Hyperparameter sweeps and ablations       &                 &                            & $\approx 200$ \\
\textbf{Project total}                    &                 &                            & $\approx 514$ \\
\bottomrule
\end{tabular}
\end{table}

\paragraph{Inference benchmarks.}
The efficiency curves in \S\ref{sec:efficiency} and Figure~\ref{fig:efficiency_scaling} were measured on the 49\,GB unit, single-batch, with 100-step warm-up and 1{,}000-step timing average per grid size. Memory was measured via \texttt{torch.cuda.max\_memory\_allocated} after a forward pass. The 11 grid sizes in Figure~\ref{fig:unified_narrative}\,(b) ($L = 100$ to $L = 22{,}500$) cumulatively required under one GPU-hour for both PSLA and softmax measurement.

\paragraph{Reproducibility note.}
DPP datasets are generated from the published DevFormer physical formula with seed 42 (App.~\ref{app:exp_dpp}); ChiPFormer uses the public \texttt{adaptec1} benchmark and ChiPFormer authors' offline trajectory dataset; CircuitNet uses the public CircuitNet 1.0 release. All training code, dataset-generation scripts, evaluation protocols, and final checkpoints will be released on acceptance.


\end{document}